\RequirePackage{fix-cm}
\documentclass[twocolumn,smallextended]{svjour3}       
\smartqed  

\usepackage{supertabular}
\usepackage[misc]{ifsym}
\usepackage{booktabs}
\usepackage{tabularx}
\usepackage{threeparttable}

\usepackage{natbib}

\usepackage{orcidlink}

\usepackage{amssymb}

\newcommand\edit[1]{#1}
\usepackage{todonotes}

{\newcommand{\nb}[2]{
		\fcolorbox{black}{yellow}{\bfseries\sffamily\scriptsize#1}
		{\sf\small$\blacktriangleright$\textit{#2}$\blacktriangleleft$}
	}
	
}

\usepackage{longtable}
\usepackage{pbox}
\usepackage{dirtytalk}
\usepackage{framed,lipsum}
\usepackage{microtype}
\usepackage{booktabs} 
\usepackage{centernot}
\usepackage{balance}
\usepackage{graphicx}
\usepackage{array}
\usepackage{color}
\usepackage{listings}
\usepackage{textcomp}
\usepackage{threeparttable}
\usepackage{centernot}

\usepackage[inline]{enumitem}
\usepackage{multicol}
\usepackage{multirow}
\usepackage{tabularx}
\usepackage{tabulary}
\usepackage{rotating} 
\usepackage[normalem]{ulem}
\usepackage{chngpage}
\usepackage{relsize}
\usepackage{xspace}
\usepackage[skins,breakable,hooks]{tcolorbox}
\usepackage{footmisc}
\usepackage{centernot}
\usepackage{wrapfig}
\usepackage{thmtools}
\usepackage{thm-restate}
\usepackage{subfig}
\usepackage[english]{babel}

\usepackage[hyphenbreaks]{breakurl}

\makeatletter
\def\cl@chapter{\@elt {theorem}}
\makeatother
\usepackage[capitalize]{cleveref}
\crefname{section}{Sec.}{sections}
\Crefname{section}{Section}{Sections}

\newcommand{\foot}[1]{\footnote{\url{#1}}}

\makeatletter
\newcommand\footnoteref[1]{\protected@xdef\@thefnmark{\ref{#1}}\@footnotemark}
\makeatother

\makeatletter
\def\tcb@restore@footnote{%
	\def\@mpfn{footnote}%
	\def\thempfn{\arabic{footnote}}%
	\let\@footnotetext\tcb@footnote@collect
}

\long\def\tcb@footnote@collect#1{%
	\expandafter\gappto\expandafter\tcb@footnote@acc\expandafter{%
		\expandafter\footnotetext\expandafter[\@thefnmark]{#1}%
	}%
}

\def\tcb@footnote@use{%
	\tcb@footnote@acc
	\global\let\tcb@footnote@acc\@empty
}
\global\let\tcb@footnote@acc\@empty

\tcbset{
	every box/.style={
		before upper pre=\tcb@restore@footnote
	},
	every box on layer 1/.append style={
		after app=\tcb@footnote@use
	}
}
\makeatother

\usepackage{array}
\newcolumntype{P}[1]{>{\centering\arraybackslash}p{#1}}

\newcounter{observationno}
\newcommand{\observation}[1]{\refstepcounter{observationno}Observation \the\numexpr\value{observationno}~(#1)}

\newboolean{showcomments}
\setboolean{showcomments}{true} 
\ifthenelse{\boolean{showcomments}}
{\newcommand{\nb}[2]{
		\fcolorbox{black}{yellow}{\bfseries\sffamily\scriptsize#1}
		{\sf\small$\blacktriangleright$\textit{#2}$\blacktriangleleft$}
	}
	
}
{\newcommand{\nb}[2]{}
	
}

\usepackage[normalem]{ulem} 
\usepackage{xcolor}

\newtcolorbox{mybox}[3][]
{
	colframe = #2!25,
	colback  = #2!10,
	coltitle = #2!20!black,
	title    = {#3},
	#1,
}

\definecolor{codegreen}{rgb}{0,0.6,0}
\definecolor{codegray}{rgb}{0.5,0.5,0.5}
\definecolor{codepurple}{rgb}{0.58,0,0.82}
\definecolor{backcolour}{rgb}{0.95,0.95,0.92}
\definecolor{xtext}{RGB}{178, 14, 140}
\lstdefinestyle{paperb}{
	commentstyle=\color{codegreen},
	keywordstyle=\color{magenta},
	numberstyle=\tiny\color{codegray},
	stringstyle=\color{codepurple},
	basicstyle=\scriptsize,
	breakatwhitespace=false,
	breaklines=true,
	captionpos=b,
	keepspaces=true,
	numbers=none,
	numbersep=5pt,
	showspaces=false,
	showstringspaces=false,
	showtabs=false,
	tabsize=2
}

\definecolor{xtextOrange}{RGB}{171,48,0}
\definecolor{xtextBlue}{RGB}{42,8,255}

\lstdefinelanguage{LTL}{
	otherkeywords={l1, l2, l3, l4, l5, chargingdock, raise_alarm, request_help, charge_battery, grasp_object, office1, office2},
	morekeywords={>,<,)},
	morecomment=[l]{.}
	sensitive=true}

\lstdefinestyle{paperc}{
	commentstyle=\color{gray},
	numbers=left,
	numbersep=5pt,
	numberstyle=\tiny\color{gray},
	keywordstyle=\bfseries\color{black},
	showspaces=false,
	showstringspaces=false,
	stringstyle=\color{xtextBlue},
	tabsize=1,
	breaklines=true
}

\lstdefinelanguage{Xtext}{
	morekeywords={grammar, with, hidden, generate, as, import, returns, current, terminal, enum},
	keywordstyle=[2]{\textbf},
	morekeywords=[3]{ID, EString, EInteger, LTL_formula, String, Integer},
	keywordstyle=[3]\color{xtextOrange},
	morecomment=[l]{//},
	morecomment=[s]{/*}{*/},
	morestring=[b]",
	morestring=[b]',
	moredelim=**[is][\color{gray}]{`}{`},
	tabsize=4
}

\begin{document}

\title{Software Reconfiguration in Robotics}

\def\makeheadbox{{%
		\hbox to0pt{\vbox{\baselineskip=10dd\hrule\hbox
				to\hsize{\vrule\kern3pt\vbox{\kern3pt
						\hbox{This work has been published in the International Journal on Empirical Software Engineering (EMSE)}
						\hbox{the published version is available at: \url{https://doi.org/10.1007/s10664-024-10596-9}}
						\kern3pt}\hfil\kern3pt\vrule}\hrule}%
					\hss}}}

\author{Sven Peldszus\textsuperscript{\orcidlink{0000-0002-2604-0487}}
	\and Davide Brugali\textsuperscript{\orcidlink{0000-0002-1024-3941}}
	\and Daniel Strüber\textsuperscript{\orcidlink{0000-0002-5969-3521}}
	\and Patrizio Pelliccione\textsuperscript{\orcidlink{0000-0002-5438-2281}}
	\and Thorsten Berger\textsuperscript{\orcidlink{0000-0002-3870-5167}}
}

\institute{
		\Letter $\;\;$  S. Peldszus \at
			Ruhr University Bochum, Germany\\
			\email{sven.peldszus@rub.de}
		\and
		D. Brugali \at
			University of Bergamo, Italy\\
			\email{davide.brugali@unibg.it}
		\and
		D. Strüber \at
			Chalmers University of Technology and the University of Gothenburg, Sweden\\
			Radboud University Nijmegen, The Netherlands\\
			\email{danstru@chalmers.se}
		\and
		P. Pelliccione \at
			Gran Sasso Science Institute (GSSI), Italy\\
			Bergen University, Norway\\
			\email{patrizio.pelliccione@gssi.it}
		\and T. Berger \at
			Ruhr University Bochum, Germany\\
			Chalmers University of Technology and the University of Gothenburg, Sweden\\
			\email{thorsten.berger@rub.de}
}
\date{}

\maketitle

\begin{abstract}
\looseness=-1
Robots often need to be reconfigurable---to customize, calibrate, or optimize robots operating in varying environments with different hardware. A particular challenge in robotics is the automated and dynamic reconfiguration to load and unload software components, as well as parameterizing them. Over the last decades, a large variety of software reconfiguration techniques has been presented in the literature, many specifically for robotics systems.
Also many robotics frameworks support reconfiguration. Unfortunately, there is a lack of empirical data on the actual use of reconfiguration techniques in real robotics projects and on their realization in robotics frameworks. To advance reconfiguration techniques and support their adoption, we need to improve our empirical understanding of them in practice.

We present a study of automated reconfiguration at runtime in the robotics domain. We determine the state-of-the art by reviewing 78 relevant publications on reconfiguration. We determine the state-of-practice by analyzing how four major robotics frameworks support reconfiguration, and how reconfiguration is realized in 48 robotics \mbox{(sub-)sys}\-tems.
We contribute a detailed analysis of the design space of reconfiguration techniques. We identify trends and research gaps.
Our results show a significant discrepancy between the state-of-the-art and the state-of-practice. While the scientific community focuses on complex structural reconfiguration, only parameter reconfiguration is widely used in practice. Our results support practitioners to realize reconfiguration in robotics systems, as well as they support researchers and tool builders to create more effective reconfiguration techniques that are adopted in practice.
\end{abstract}




	\section{Introduction}
	\label{sec:intro}
	\noindent
	\looseness=-1
	Robots are increasingly deployed in our lives. Being multi-purpose, they can operate in a variety of safety-critical environments, such as private homes, hospitals, restaurants, factories, and museums.
	Such robots have a variety of mobility and manipulation devices, as well as redundant sensors brought together in a robotic control system---``\emph{an interconnection of components forming a system configuration that will provide a desired system response}'' \citep{2011Dorf}.

	Robots are concurrent and distributed software systems \citep{1996Aarsten} composed of many components.
Configuring and assembling these components---many with alternative implementations---is essential to customize robotic systems towards different hardware, runtime environments, or non-functional properties (e.g., performance or energy consumption).
	In fact, the initial release of the Robot Operating System (ROS) \citep{ram2010ROS} in 2010 already contained hundreds of open-source components stored in 15 repositories, accounting for customization needs and enhancing robot versatility to changing application requirements.
	To this end, so-called configuration mechanisms enabling customization are considered essential by robot manufacturers and developers \citep{garcia2020robotics,GarciaEMSE2022}.

	\looseness=-1
	\edit{\emph{Configuration determines which components of a system are present and active,
		and how they are connected. It also instantiates the component parameters.}}
	Configuration is often not static, but \emph{reconfiguration} is typically needed to assure the correct and trustworthy operation of robotic systems.
	\edit{Reconfiguration \citep{kortenkamp2016robotic} often involves the activation or deactivation of components, the modification of their connections, the replacement of control and functional algorithms, and changing control parameters at runtime.} \edit{Also, reconfiguration needs to be triggered, which is usually provided via specific interfaces}.
\edit{The topic of reconfiguration is quite broad. This article focuses on automated software reconfiguration at runtime in robotics control systems.
	Static configuration and manual reconfiguration, but also related topics under the umbrella of self-adaptation, are beyond the scope of this article.}

\edit{
	\emph{Self-adaptation} is the ability of a system to dynamically adapt to unexpected environmental changes and failures.
	Self-adaptation subsumes reconfiguration.
	It allows a system to adapt its configuration when the conditions change, to deal with uncertainties that are difficult or impossible to anticipate before deployment \citep{Calinescu2020}.
	Uncertainty is a system state of incomplete or inconsistent knowledge caused, e.g., by unpredictable phenomena in the execution environment or incomplete and inconsistent information obtained by potentially imprecise, inaccurate, and unreliable sensors 
	 \citep{Cheng2012}.
	At runtime, a self-adaptive system collects additional information to resolve the uncertainty and adapt itself \citep{Calinescu2020}.
	To this end it may use reconfiguration.
	For example,  let us assume that a configuration parameter is the nominal maximum speed of a robot, which depends on the robot kinematics.
	Then, the actual maximum speed of the robot is adapted (i.e., reduced) by using reconfiguration for meeting safety requirements, taking also environment uncertainty (e.g., wet floor) into account.
	When the robot is out of safety-critical zones or conditions, the maximum speed is adapted (i.e., increased).
	Finally, the actual speed is continuously regulated according to the path geometry.
	As an alternative to reconfiguration, self-adaptation may also use other techniques, i.e., internal variables, as an alternative to configuration parameters to achieve the intended behavior.
}

\looseness=-1
	To enable effective and safe reconfiguration as described above, specific implementation techniques are needed.
	To this end, the scientific community has conducted extensive research on software reconfiguration of robotic systems.
	At the same time, popular robotics frameworks, such as the Robotic Operation System (ROS) \citep{quigley2009ros}, YARP \citep{Metta2006,Fitzpatrick2008}, or Smartsoft \citep{smartsoft_project} provide mechanisms for implementing reconfiguration.
	However, surprisingly little is known about the adoption and characteristics of reconfiguration mechanisms in practice.
	While many secondary studies on the physical reconfiguration of modular robots and robot swarms (see \cref{sec:background}) exist,
	to the best of our knowledge, there is no systematic overview of available technologies for designing reconfigurable robotic systems. 
	To improve the situation, \textit{the state of the art in robot software reconfiguration needs to be assessed}---the first motivation for our study.

\looseness=-1
Reconfiguration can be even more difficult in practice. Static configuration already adds substantial complexity to the design of a software system~\citep{berger2013study}, where reusable software components are selected and integrated during deployment according to the specific requirements of the robot embodiment, task, and environment. Empowering the robot with the ability to dynamically self-reconfigure at runtime according to context changes requires to face additional challenges.
For instance, developers need to (i) define triggers of reconfiguration, (ii) declare constraints among the configuration options, consistent with the domain knowledge and the actually implemented system, and (iii) assure that reconfiguration is not only timely, but also puts the system into the desired, and valid state of operation.
To improve the situation, \textit{the state of the practice in robot software reconfiguration needs to be assessed}---the second motivation for our study.

	It can be argued that tool and framework builders should enhance the reconfiguration mechanisms to benefit practitioners, who could then build more reliable and safe reconfigurable robotics software \citep{Dalal1993}.
	However, to the best of our knowledge, there are no studies or at least experience reports that address the practical implementation of dynamic reconfiguration.
	This lack of empirical data impedes the development of effective reconfiguration methods and tools, as well as the scoping of research projects and the selection of relevant research directions.

\looseness=-1

To address these shortcomings, we present a comprehensive review of the literature on reconfiguration of robotic software systems (state-of-the-art) and an in-depth analysis of reconfiguration support in robotic frameworks, as well as the implementation of reconfiguration in robotic (sub-)systems (state-of-practice).
Our research questions are:

\vspace{4pt}
\noindent
{\bf RQ1}: {\em What are the motivations for developing dynamically reconfigurable robotic software systems?}

We aim at understanding the needs of reconfiguration in robotics, e.g., for enhancing performance, robustness, and reusability.
To this end, we analyze the literature on robotic software reconfiguration.

\vspace{4pt}
\noindent
{\bf RQ2}: {\em What software parts of a reconfigurable robot control system are reconfigured and at what granularity?}

When talking about reconfiguration in robotics, it is often not clear which parts of a robotic system are adapted to needs using reconfiguration or maybe other mechanisms.
We aim at understanding what is commonly considered in reconfiguration and at which granularity, e.g., \edit{entire components or single software functionalities.}
To this end, we address this question from two perspectives, first by analyzing the literature on robotic software reconfiguration, and second by also reviewing robotic frameworks in terms of the granularity of reconfiguration that they support.

\vspace{4pt}
\noindent
{\bf RQ3}: {\em What mechanisms exist and how they are used for developing reconfigurable robotic software systems?}

We aim at understanding the mechanisms used for specifying and performing the reconfiguration, together
with implementation practices.
To this end, we analyze the literature on robotic software reconfiguration, we review robotic frameworks and their support for reconfiguration, and we analyze how researchers and practitioners implement reconfiguration in their robotic (sub-)systems to answer RQ3.

\looseness=-1
Based on the academic literature, robotic frameworks, and robotic \mbox{(sub-)sys}\-tem implementations, we address reconfiguration from two perspectives: (i) an academic perspective based on our systematic literature review (SLR) and (ii) a practitioner's perspective based on how robotic frameworks address reconfiguration and how reconfiguration is implemented in real robotic \mbox{(sub-)sys}\-tems.
We synthesize the individual perspectives to identify discrepancies and common perceptions of software reconfiguration in robotics, thereby providing a comprehensive view of reconfiguration.
Additionally, we contrast the state of practice in reconfiguration of robotic systems with reconfiguration in other domains and derive implications for researchers and practitioners.
	\section{Background and Related Work}
	\label{sec:background}
	\noindent
	In this section, we first introduce background and related work on reconfiguration in robotics, and thereafter, an overview of six robotic frameworks.

\subsection{Overview of Reconfiguration in Robotics}
\label{sec:back:reconf}
A first initial search for systematic literature reviews and surveys on reconfiguration of robotic systems (via ACM Digital Library, Scopus, and IEEE Explore) by March 2024 revealed that the terms \textit{``reconfigurability''} or \textit{``reconfigurable''} or \textit{``reconfiguration''} have been used in the robotics literature with different meanings depending on the type of robotic system.

The common ground on which all studies agree is that reconfigurability implies an ease of modification and an absence of irreversible or rigid commitments in some aspects of the robotic system (mostly related to its embodiment) and the potential to assume different arrangements of the constituent elements.

In the last 30 years, surveys have been published on three classes of reconfigurable systems.
Robots composed of multiple identical modules can be mechanically assembled to form systems with different physical shapes (e.g., a snake robot or a spider robot) and can autonomously reconfigure their shape for adapting to different environments \citep{Dudek1993,Yim2007,Moubarak2012,Ahmadzadeh2015,Chennareddy2017,Alattas2018}.
Robots composed of modules with specialized functionalities can self-reconfigure to perform different tasks \citep{Liu2016}.
Swarms of mobile robots (e.g., swarms of UAVs) that can reconfigure the swarm topology resulting from coalition formation during the execution of cooperative tasks \citep{Abukhalil2015,Shlyakhov2017}.

The Robotics 2020 Multi-Annual Roadmap (MAR) \citep{MAR} provides a more general definition of configurability, as ``\emph{the ability of the robot to be configured to perform a task or reconfigured to perform different tasks.}'' This may range from the ability to re-program software modules and components to being able to alter the configuration of sensing and other electronic systems and the mechanical structures of the system.

Moreover, the MAR distinguishes reconfiguration from the concept of adaptability, which is defined as ``\emph{the ability of the system to adapt itself to different work scenarios, different environments and conditions}'' and implies that the system performs optimization against some performance criteria.


Interestingly, two secondary studies investigated the intriguing relationship between reconfigurability and adaptability and are specifically relevant for our investigation because they introduce the concept of dynamic reconfiguration.
More specifically, \cite{SLRReconfigurable} define a Dynamic Reconfigurable System as ``\emph{a system whose subsystems can be modified or have their configurations changed during operation; dynamic reconfiguration enables real-time systems adaptation.}'' The paper focuses on two specific aspects: the reconfiguration of the computing hardware (mostly FPGAs) and, although only marginally, the reconfiguration of the robot control system (e.g., the autonomous navigation system).
\cite{Tan2020} define self-reconfigurable robots as ``\emph{machines that can change their morphologies as per prescribed requirement or are adaptable to the environments with provided level of autonomy.}''
In turn, autonomous reconfigurability is described as ``\emph{the extent to which a self-reconfigurable robot can sense its environment, plan its configuration based on that environment, and act to transform into specific configurations upon that environment with the intent of achieving some goal.}''

In the context of this paper, we focus on \edit{automated runtime reconfiguration} of the software control system of autonomous robots.
In order to clarify the scope of the paper, it is useful to refer to concepts and definitions that we have found in some works included in our study.

\cite{Stewart1997} argue that the need for dynamic reconfiguration stems from the need to change control algorithms on the fly to support more intelligent control~strategies.

\cite{Jamshidi2019} propose a technique to enable self-adaptation of robots operating in dynamic and uncertain environments, using configuration change as the main mechanism to enact adaptation. In their paper, robotics software is considered as a highly-configurable system, in which system characteristics (e.g., usage of sensors) are treated as configuration options.

\cite{Macdonald2004} explain that robot software reconfiguration requires specific mechanisms for component deployment. In some cases, state information (e.g., map data or sensor data history) must be transferred to a newly deployed software component during robot operation without service disruption.

\cite{Pham2000} discuss three forms of software adaptation in robotic systems: parametric fine tuning, algorithmic change, and task migration to remote computational resources in a distributed environment.


\subsection{Robotic Frameworks}
\label{sec:back:frameworks}
During the last twenty years, several research and development projects have produced a variety of component-based robotic frameworks (see the surveys  of \cite{2007BrugaliTrendsFw} and \cite{2012RoboticMidlewareSurvey}) that in many cases build on state-of-the-art middleware infrastructures (e.g., CORBA \cite{OMG_CORBA} and DDS \citep{OMG_DDS}).
These frameworks provide domain-specific software abstractions that are amenable to robotics experts and hide the complexity of middleware mechanisms for real-time execution of concurrent control activities, synchronous and asynchronous communication between components, dynamic wiring of component interfaces, remote configuration of component properties, and runtime loading of plug-in functionality.
In the following, we briefly introduce some of these frameworks.

GenoM \citep{1997Genome} is a component-oriented software framework developed by the LAAS CNRS robotics group. Components interface hardware devices or encapsulate common robotics algorithms. GenoM components are collections of control services, which manage incoming requests, implement specific algorithms, and execute services. The framework is at the basis of the LAAS Architecture \citep{1998LAAS}, one of the most relevant examples of robot architectures that enforces software quality attributes, such as modularity, maintainability, and usability.

The Claraty \citep{2001CLARAty} framework has been developed with the specific goal of improving the modularity, interoperability, and reusability of the control software operating the large variety of NASA/JPL autonomous robots for planetary exploration. The framework provides mechanisms for encapsulating perception capabilities, robot action, and control loops that are encapsulated into software components, which can be activated by the decisional level.

\emph{OROCOS} is one of the oldest open-source frameworks in robotics, under development since 2001.
Professional industrial applications and products use it since 2005.
Its focus has always been to provide a hard real-time component framework, the so-called Real-Time Toolkit (RTT), that is as independent as possible from any communication middleware and operating system.

\emph{YARP}~\citep{Metta2006,Fitzpatrick2008} is a multiplatform and multiprotocol communication framework for robotic research.
Available protocols are tcp, udp, multicast, shared memory, and protocols for interfacing with ROS.
Typical YARP applications consist of several intercommunicating modules distributed on different machines that exchange messages according to a port-based publisher/subscriber protocol.
YARP is the reference software platform for the iCub humanoid robot designed by the Italian Institute of Technology to help developing and testing embodied AI algorithms. The iCub robot is currently used by more than thirty research institutions worldwide.

\emph{RobMoSys}~\citep{robmosys} is a EU-funded research project that provides interfaces, methods, and tools for the model-driven development of robotic systems.
It enables the management of interfaces between different robotics-related domains, roles, and levels of abstractions.
Its interfaces can either be implemented by concrete frameworks or wrappers around low-level implementations mapping them to the RobMoSys APIs.
For our study on software reconfiguration, we consider the SmartSoft framework~\citep{schlegel1999software,Schlegel2013} as a reference implementation of RobMoSys.
While the \emph{RobMoSys} ecosystem, like many tools developed by academia, suffers the problem of sustainability after the completion of the project, the SmartSoft framework has reached the Technology Readiness Level (TRL) 6 and is used in industrial projects in collaboration with several companies, including Bosch, REC, and FESTO.


\emph{ROS}~\citep{ros} is a mature open-source robotics middleware that provides a framework for robotic software.
It allows implementing modular robotic applications in multiple programming languages and provides services
to realize the interaction of modules.
Many state-of-the-art algorithms have been developed for ROS, and major robotic systems, such as the self-driving vehicle software Autoware.auto, are implemented using ROS.
The transition from ROS to ROS 2~\citep{2022Macenski} was accompanied by the definition of a more standard structure of ROS packages (e.g., the navigation stack), which use behavior trees to orchestrate the behavior of many concurrent functionality and their runtime reconfiguration. What is still missing are reference architectures that guide application developers in the integration of ROS packages in whole applications.

\cite{Malavolta2021} mined robotics software repositories to extract guidelines for building robotic systems.
Although the extracted guidelines do not explicitly address reconfiguration, they cover aspects necessary for building reconfigurable systems.
First, there are guidelines that focus on increasing component cohesion and reducing component coupling, which are essential for reconfiguration. Second, extracted design guidelines focus on components that provide interfaces to adjust their operation, such as changing the frequency at which sensors collect data. Third, guidelines focus on providing status and health information, which can be seen as potential triggers for reconfiguration. Finally, some guidelines focus on separating the monitoring of such health information from the functional implementations.
\section{Methodology} \label{sec:methodologyro}
%

\begin{figure}[t]
	\begin{center}
		\includegraphics[width=\columnwidth]{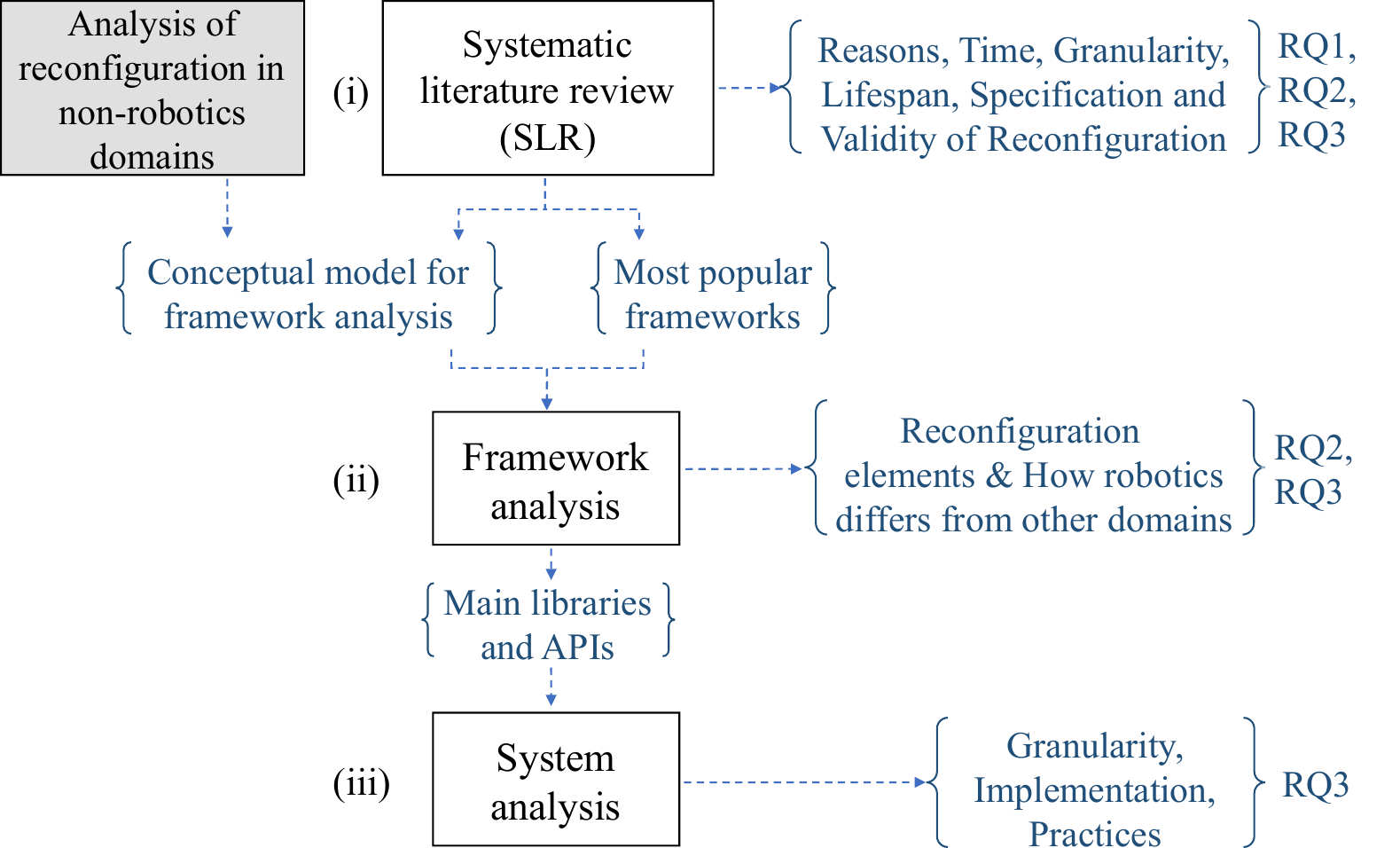}
		\caption{Overview on our research methodology}
		\label{fig:overview}
	\end{center}
\end{figure}

\noindent\looseness=-1
Since we want to study how \edit{automated} robot reconfiguration \edit{at runtime} is addressed by the scientific community and contrast it with how it is implemented by practitioners, we decided to study three different types of artifacts to answer the research questions.
As illustrated in \cref{fig:overview}, we studied the state-of-the-art and the state-of-practice on reconfiguration in the robotics domain in three phases. Thereby, we applied two research methodologies (systematic reviews and repository mining), as described by \cite{emp-std}, depending on the investigated artifacts:

\begin{enumerate}[label={(\roman*)}]
	\item \textbf{Literature:} \emph{Systematic Review} of the literature on the \edit{automated} reconfiguration of robotic software systems \edit{at runtime}.
	\item \textbf{Frameworks:} \emph{Systematic Review} of robotic frameworks and their support for reconfiguration.
	\item \textbf{Implementations:} \emph{Repository Mining} to analyze how reconfiguration is implemented in open-source robotic (sub-)systems.
\end{enumerate}

Specifically, we conducted the systematic literature review (SLR) to capture the scientific perspective on \edit{automated software reconfiguration at runtime} in robotics and to extract theoretical foundations for the subsequent studies.
We focused at identifying the reasons for reconfiguration, the expected lifespan of a configuration, and the granularity of reconfiguration, along with the means for specifying reconfiguration.
Based on the robotic frameworks mentioned in the papers surveyed in the SLR and additional literature, we identified the most used frameworks for analysis in phase (ii) and which ones to investigate in phase (iii).

Taking into account the observations from the SLR, we synthesized a conceptual model as a basis for systematically analyzing robotic frameworks.
Conceptual models are means for systematically capturing and communicating common knowledge about a domain \citep{Gemino2004}, therefore, being a suitable means to build a common basis according to which we can classify and compare robotics frameworks.
Based on the identified conceptual model, we then analyzed and classified robotics frameworks in phase (ii).

Our goal in phase (iii) is to determine what role \edit{automated} reconfiguration plays in the software of robotic \mbox{(sub-)systems} and what practices are used to implement it.
To enable this analysis of concrete implementations of reconfiguration in robotic (sub-)systems (iii), during the analysis of robotic frameworks (ii), we also focused on identifying the main libraries used in robotics that can be used to implement reconfiguration.

Our replication package provides all raw data, scripts that have been used, and results~\citep{replication}.\footnote{Replication package at Zenodo: \url{https://doi.org/10.5281/zenodo.14013818}}

\subsection{Systematic Literature Review}
The goal of our SLR is to capture the academic perspective on reconfiguration in robotics and what techniques exist to address reconfiguration.
While we do not expect the academic perspective to perfectly match practice, we do expect it to at least cover the practical aspects, but also to provide a broader range of techniques and considerations that may not yet be practical.

\begin{figure*}[t]
	\center
	\includegraphics[width=.8\textwidth]{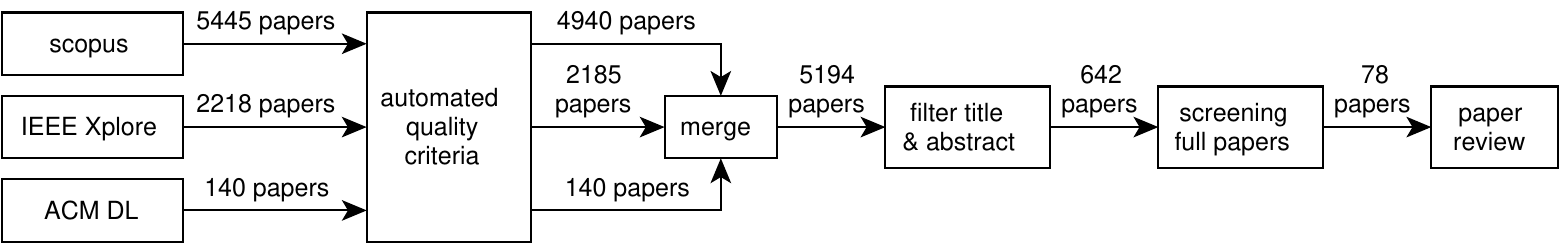}
	\caption{Selection process of the primary studies}
	\label{fig:selection}
\end{figure*}

As described in \cref{sec:background}, most publications dealing with robot reconfiguration mainly focus on reconfigurable hardware whereas
software aspects related to the reconfiguration of robots are not treated as main concerns.  
Nevertheless, the software configuration of a robot control system is highly affected by its mechanical structure, task, and operating environment \citep{GarciaEMSE2022}.
While some papers (e.g., \cite{Stueben2021}) present case studies that motivate the runtime reconfiguration of the robotic systems, they do not give a systematic overview.
For this reason, we defined a generic search string based on only three keywords:

\begin{center}
	\texttt{("reconfigur*" OR "re-configur*") AND (robot*)}
\end{center}

The search string was customized for the IEEE Xplore, Scopus, and ACM DL and applied to the title and abstract.
Following the process shown in \cref{fig:selection}, the results were automatically filtered according to quality criteria to exclude studies not written in English, short papers (\textless{} 3 pages), posters, and workshop summaries.

Even though the keyword search was limited to the fields ``Title'' and ``Abstract,'' it resulted in a high number of studies. The searches returned 2,185 results on IEEE Xplore, 4,940 results on Scopus, and  140 results on the ACM DL. We implemented a script to merge the results and remove the duplicates, which returned a total of 5,194 results.

We aimed at analyzing how the software reconfiguration of robots is implemented. Consequently, we selected those papers that fulfill the following inclusion criteria:

\begin{itemize}
	\item[I1:] Studies focusing on the software of robotic systems;
	\item[I2:] Studies that present examples of robots that perform complex tasks, i.e., that require the integration and configuration of a variety of functionalities (e.g., perception, control, planning);
	\item[I3:] Studies that consider changing the configuration of robotic systems \edit{at runtime}.
\end{itemize}

\noindent
In addition, we applied the following exclusion criteria to exclude works that primarily focus on parts other than the software of a robotic system, or robotic systems that do not themselves include significantly complex robotic control software:

\begin{itemize}
\item[E1:] Studies focusing on single-purpose robotic demonstrators, microrobots, and robotic devices (e.g., a robot hand), since they controlled from the outside or do not run software on their own, and therefore, their software cannot be reconfigured directly;
\item[E2:] Studies that do not deal with robot control (e.g., papers solely focusing on reconfigurable computer architectures and communication networks, the mechanical design of metamorphic robotic systems);
\item[E3:] Studies that focus only on algorithms (e.g., for coalition formation of multi-robot systems or kinematic calibration of configurable robots) and do not include reconfiguration of the individual robots involved; 
\item[E4:] The software part of reconfiguration plays only a minor role and other aspects such as hardware reconfiguration are the main focus.
\end{itemize}


A significant portion of the 5,194 papers that resulted from the high yield and low precision associated with our generic search string 
 was filtered out by scanning the title and abstract. This step was performed by the second author with 20 years of experience in robotics research at a reading rate of three studies per minute.
This step identified 642 full papers.
Thereafter, we applied the inclusion and exclusion criteria based on the introduction and conclusion of the paper, \edit{resulting in 78 primary studies} for analysis.

These papers were then randomly divided into three groups for review and classification of the full papers.
Each group was then classified by a different author according to criteria relevant to reconfiguration.
If the classification of a paper was in question regarding one of the criteria, it was analyzed by at least three authors.
Disagreements were resolved by involving all authors, who discussed and reached agreement.
We carried out this step over a period of 4 weeks, with a weekly meeting where we discussed between two and four concrete classifications per meeting.

During this full paper review, we excluded \edit{46} of the primary studies based on the content of the full paper.
This exclusion was because they were duplicates of other included papers, e.g., a paper is also included as an extended journal version (6 papers), or the content did not include a contribution to \edit{automated software reconfiguration at runtime}, although we had first expected that based on the introduction and conclusion \edit{(42 papers)}.

Based on the primary studies, we performed then one iteration of backward snowballing, resulting in 1,040 papers cited by the primary studies.
As for the primary studies, we again filtered these papers based on title and abstract using the exclusion criteria from above, resulting in 125 potentially relevant papers.
These 125 potentially relevant papers were then filtered based on the full paper, resulting in an additional \edit{21 papers} for analysis.
As before, in each step of the snowballing process, we randomly divided the papers into three groups, each of which was reviewed by a different author.
Because of our experience with the classification of the primary studies, we did not need to have group discussions for the remaining papers.

Overall, the review process resulted in a total of \edit{78 papers} that were analyzed in detail to answer the research questions.
\Cref{tab:papers-initial} gives an overview of the papers identified in the initial search, and \cref{tab:papers-bwd} of the papers identified via backward snowballing.

\begin{table}
	\caption{Papers identified in the SLR's initial search}
	\label{tab:papers-initial}
	\center
	\scriptsize
	\begin{tabular}{ll}
		\toprule
		\textbf{ID} & \textbf{Paper} \\ 
		\midrule
		P1	& \cite{BergeCherfaoui1994}\\ 
		P2	& \cite{HayesRoth1995}\\ 
		P3	& \cite{Schneider1995}\\ 
		P4	& \cite{Stewart1997}\\ 
		P5 	& \cite{Budenske1997}\\ 
		P6	& \cite{Vos1998}\\ 
		P7	& \cite{Fayman1998}\\ 
		P8	& \cite{Boluda1999}\\ 
		P9	& \cite{Gafni1999}\\ 
		P10	& \cite{Lindstroem2000}\\ 
		P11	& \cite{Karuppiah2001}\\ 
		P12	& \cite{Kubota2001}\\ 
		P13	& \cite{Zhang2001}\\ 
		P14	& \cite{Cobleigh2002}\\ 
		P15	& \cite{Kim2003}\\ 
		P16	& \cite{Bi2003}\\ 
		P17	& \cite{Inohira2003}\\ 
		P18	& \cite{Kim2004}\\ 
		P19	& \cite{Roh2004}\\ 
		P20	& \cite{Lee2005}\\ 
		P21	& \cite{Lee2006}\\ 
		P22	& \cite{Kim2006}\\ 
		P23	& \cite{Yu2006}\\ 
		P24	& \cite{Maeda2006}\\ 
		P25	& \cite{Kim2006a}\\ 
		P26	& \cite{Brandstoetter2007}\\ 
		P27	& \cite{Scheutz2007}\\ 
		P28	& \cite{Braman2007}\\ 
		P29	& \cite{Morris2007}\\ 
		P30	& \cite{Cabrol2008}\\ 
		P31	& \cite{Lee2008a}\\ 
		P32	& \cite{Nilsson2008}\\ 
		P33	& \cite{Lee2008b}\\ 
		P34	& \cite{Santos2009}\\ 
		P35	& \cite{Xiao2012}\\ 
		P36	& \cite{Benmoussa2013}\\ 
		P37	& \cite{Marques2013}\\ 
		P38	& \cite{Goldhoorn2014}\\ 
		P39	& \cite{Scala2014}\\ 
		P40	& \cite{Szlenk2015}\\ 
		P41	& \cite{Frost2015}\\ 
		P42	& \cite{Shaukat2016}\\ 
		P43	& \cite{Meszaros2017}\\ 
		P44	& \cite{Doose2017}\\ 
		P45	& \cite{SLR-2018-Brugali-2}\\ 
		P46	& \cite{Jamshidi2019}\\ 
		P47	& \cite{Ramachandran2019}\\ 
		P48	& \cite{Murwantara2020}\\ 
		P49	& \cite{Cardoso2019}\\ 
		P50	& \cite{Cruz2020}\\ 
		P51	& \cite{2020BrugaliROS}\\ 
		P52	& \cite{camara2020software}\\ 
		P53	& \cite{Bozhinoski2021}\\ 
		P54	& \cite{Pane2021}\\ 
		P55	& \cite{Stueben2021}\\ 
		P56	& \cite{Kozov2021}\\ 
		P57	& \cite{Nordmann2021}\\ 
		\bottomrule
	\end{tabular}
\end{table}

\begin{table}
\caption{Papers identified in the backward snowballing of the SLR}
\label{tab:papers-bwd}
\scriptsize
\center
\begin{tabular}{ll}
	\toprule
	\textbf{ID} & \textbf{Paper} \\ 
	\midrule
		P58	& \cite{Ferrell1994} \\ 		
		P59	& \cite{Lee1994} \\ 			
		P60	& \cite{Firby1995} \\ 			
		P61	& \cite{Stewart1996} \\			
		P62	& \cite{Pham2000} \\			
		P63	& \cite{Wills2001} \\			
		P64	& \cite{Macdonald2004} \\		
		P65	& \cite{Kramer2006} \\			
		P66	& \cite{Parker2006} \\			
		P67	& \cite{Yoo2006} \\				
		P68	& \cite{Calisi2008} \\			
		P69	& \cite{Edwards2009} \\			
		P70	& \cite{Tajalli2010} \\			
		P71	& \cite{francisco2012dealing} \\
		P72	& \cite{Hartmann2013} \\		
		P73	& \cite{Iftikhar2014} \\		
		P74	& \cite{7353608} \\				
		P75	& \cite{Leng2016} \\			
		P76	& \cite{Aguado2021} \\			
		P77	& \cite{Nordmann2021} \\		
		P78	& \cite{Bozhinoski2022} \\		
		\bottomrule
	\end{tabular}
\end{table}



\subsection{Robotic Frameworks Review}
\label{sec:method-frameworks}

%

We selected and analyzed common frameworks used in robotics to identify at what granularity reconfiguration is technically supported (RQ2) and what interfaces frameworks provide to robotics software developers that can be used for implementing reconfiguration (RQ3).
For all frameworks, we investigated how configurable software parts can be \emph{specified} and \emph{encapsulated}, how these can \emph{interact} with each other, and how reconfiguration is \emph{planned} and \emph{executed}.


\subsubsection{Framework Selection}
To increase the relevance of the frameworks selected for the review, we considered observations from the SLR in the selection process.
In our SLR, 24 different frameworks were mentioned but only Chimera (a robot programming environment from the 90s~\citep{Stewart1990}), real-time CORBA (an OMG standard for real-time management of objects from the 2000s~\citep{Schmidt2000}), and ROS~\citep{ros} were mentioned more than once.
The first two were not mentioned in papers after 2006.
For this reason and for their age, we considered them as dated and, did not investigate them in depth.
Consequently, besides ROS, we selected alternatively the most popular robotic frameworks based on existing surveys on software engineering practices in the service robotics domain~\citep{GSB2020,garcia2020robotics}.
%
ROS is used by 88.5\% of the survey participants, followed by its successor ROS2 (22.4\%)
and  OROCOS (18.6\%).
After a larger gap in popularity, YARP and SmartSoft follow with 4.5\% and 3.8\%.
As SmartSoft is a collection of concepts and tools for RobMoSys~\citep{smartsoft_project}, an abstract framework for the model-based development of robotic systems~\citep{robmosys}, we decided to consider SmartSoft as one example for a practical realization of RobMoSys.

In summary, we selected the following frameworks for an analysis:
%
(i) ROS (and ROS2),
(ii) OROCOS,
(iii) YARP, and
(iv) RobMoSys (incl. SmartSoft).
%

	\subsubsection{Conceptual Model}
	\looseness=-1
	The first step in actually investigating the robotics frameworks is to identify the concepts we will use to investigate the frameworks.
	To this end, to systematically derive the relevant concepts and ensure their completeness, we have created a conceptual model for software reconfiguration in robotics.
	This conceptual model will serve as a means to identify relevant concepts that frameworks must support to enable reconfiguration and to systematically analyze the frameworks concerning them.
	We proceeded in the creation of the conceptual model as follows.

We started the synthesis of the conceptual model from an existing schema \citep{berger.ea:2014:ecosystems}, which was proposed in a study on variability mechanisms in software ecosystem platforms (e.g., Linux kernel, Eclipse plugins, and Android apps), and adapted it to the robotics domain based on insights from the SLR and further literature \citep{Mens2003,Fritsch2008,Eddin2013,Krupitzer2015,Mens2016,Tan2020}.
Therefore, we first compared this model with our observations on the scientific understanding of reconfiguration in robotics and excluded aspects that are not applicable from this initial model.
Thereafter, we added further concepts that we observed in the SLR based on the additional literature.

\subsubsection{Review Process}
Using the derived conceptual model, we classified the robotic frameworks based on their documentation and scientific publications about them.
The first and second authors of this paper independently read the documentation provided for the robotic frameworks, mainly wikis, the websites of the robotic frameworks, and related publications, and recorded relevant references for each part of the conceptual model. \Cref{tab:frameworks-sources} lists the sources investigated during the frameworks review. For the identification of the features of these frameworks that are relevant for our investigation, we also analyzed tutorials (e.g., by \cite{brugali2009component_1} and \cite{brugali2009component_2}), experience reports (e.g., by \cite{2020BrugaliROS}), and secondary studies (e.g., by \cite{ahmad2016software} and \cite{2023Albonico}).

Thereafter, the two authors discussed and consolidated their findings.
For aspects for which no agreement was achieved, the authors again independently investigated the available resources and then continued with the consolidation.
Full agreement was achieved after four iterations.
Based on the classifications, we then compared the robotic frameworks with each other and also with the findings from the SLR.

\begin{table*}
	\caption{Data sources used during the frameworks review}
	\vspace{-.5cm}
	\label{tab:frameworks-sources}
	\center
	\scriptsize
	\def\arraystretch{1.25}
	\begin{tabular}{rllll}
		\toprule
			& \textbf{Wikis} & \textbf{Websites} & \textbf{Publications} \\
		\midrule
		ROS (ROS2) & \href{https://wiki.ros.org}{wiki.ros.org}	& \href{https://www.ros.org}{www.ros.org}	& \cite{quigley2009ros,estefo2019robot}; \\
		 & \href{https://docs.ros.org}{docs.ros.org}	& 	& \cite{ram2010ROS,whyROS2} \\[+.1cm]
		OROCOS & \href{https://docs.orocos.org}{docs.orocos.org}	& \href{http://www.orocos.org}{orocos.org}	& \cite{2001orocos} \\[+.1cm]
		YARP & \href{https://wiki.icub.eu/wiki/YARP}{wiki.icub.eu/wiki/YARP}	& \href{https://www.yarp.it}{yarp.it}	& \cite{Metta2006} \cite{2015Paikan} \\[+.1cm]
		RobMoSys 	& \href{https://robmosys.eu/wiki}{robmosys.eu/wiki}	& \href{https://robmosys.eu}{robmosys.eu}	& \cite{lotz2013managing} \\
			(SmartSoft)					& \href{https://wiki.servicerobotik-ulm.de}{wiki.servicerobotik-ulm.de}	&  \href{https://smart-robotics.sourceforge.net}{smart-robotics.sourceforge.net} & \cite{schlegel1999software} \cite{2021Schlegel}\\
		\bottomrule
	\end{tabular}
\end{table*}

\subsection{Investigation of Robotic (Sub-)Systems}
\looseness=-1
To identify what mechanisms are used in practice and how they are used (RQ3), we mined data from robotic software repositories.
In doing so, we focused on robotics (sub-)systems that contain usages of the interfaces of robotic frameworks that we identified in the frameworks investigation as being suitable to implement reconfiguration although this might not be their main purpose.
We realized that among the identified robotic frameworks, only ROS~\citep{quigley2009ros} is mature enough to serve as a convincing basis for the in depth evaluation of robotic (sub-)system implementations.
This is also supported by the findings from our SLR and the survey of \cite{garcia2020robotics}.

To select suitable ROS (sub-)systems, we use a list of open-source ROS (sub-)systems repositories that has been created by Malavolta et al. in 2021 when studying general guidelines for developing robotic \mbox{(sub-)}sys\-tems \citep{Malavolta2021}.
They define a ROS-based system as ``\emph{a system that contains robotic capabilities built using the ROS framework}.''
To identify suitable ROS-based systems, Malavolta et al. mined GitHub, GitLab, and Bitbucket and did extensive quality assessments to identify deployable ROS (sub-)systems that implement huge parts of their logic on their own and are not only wrappers for libraries or toy examples.
Thereby, implementing the logic of the system does not prevent it from using libraries such as the ROS navigation stack\footnote{ROS Nav2: \url{https://navigation.ros.org/}} to realizing this logic.

To systematically investigate how reconfiguration is implemented, we use the APIs of ROS that we identified in the frameworks review as being suitable to implement reconfiguration as entry points for our software inspections.
Starting from these entry points, we inspected how the API is used to understand how reconfiguration is implemented in the robotic (sub-)systems.
Since, many APIs allowed to register callback functions, we followed the calls in both directions forward and backwards to  visit all relevant code parts.
\section{Conceptual Model of Reconfiguration}
\label{sec:model}

\begin{figure}[b]
	\includegraphics[width=\columnwidth]{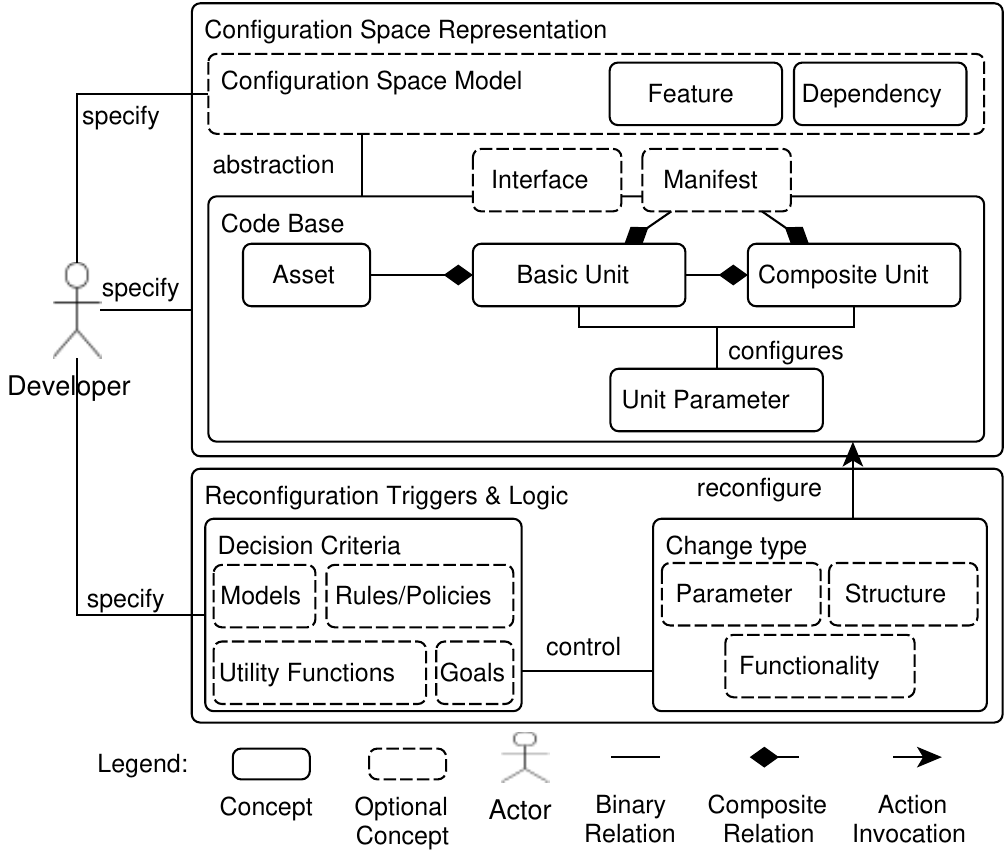}
	\caption{Conceptual model of reconfiguration in robotics}
	\label{fig:template}
\end{figure}

For investigating robotic frameworks, we created a conceptual model of reconfiguration in robotics according to which we classified the investigated robotic frameworks.
\Cref{fig:template} shows the identified concepts as well as their relations. 
Although this conceptual model is partially based on findings from our SLR, we present it here prior to the detailed discussion of the findings of that SLR to allow for a discussion of findings from all three types of artifacts reviewed together.

We started to create the conceptual model for reconfiguration in robotics from an existing conceptual model proposed in a study on variability mechanisms in software ecosystem platforms~\citep{berger.ea:2014:ecosystems}.
Then, based on the insights from the performed SLR, we excluded all aspects that do not apply to reconfiguration in robotic systems.
For example,
	we conclude that in robotics, decisions to bind a feature as active or inactive are only temporal and made dynamically.
For this reason, we excluded the decision lifecycle and binding-related aspects from our framework.
As we focus on open-source robotic frameworks, we also dropped the aspect of platform openness.

While the reduced conceptual model captures what can be reconfigured, how reconfigurable assets are specified, and how these can interact, it does not capture how reconfiguration is specified and executed, yet.
In our SLR we have seen that decision making is an essential part of reconfiguration, and there are various ways of specifying it.
In an investigation of self-adaptation in robotic systems~\citep{Krupitzer2015}, such specifications have been categorized into four kinds of decision criteria for specifying reconfiguration triggers and actions (Models, Rules/Policies, Utility Functions, and Goals), introduced in detail in \cref{sec:trigger}.
To use the same naming as the existing literature, we added the decision criteria to our conceptual model.
Also, we identified \edit{three} fundamental change types realized by reconfiguration mechanisms in the literature~\citep{Fritsch2008,Krupitzer2015}, namely changing parameter values, software structure, and functionality, which we introduce in detail in \cref{sec:trigger}.
Thereafter, we iteratively confirmed and adjusted our conceptual model based on additional related literature~\citep{Mens2003,Fritsch2008,Eddin2013,Krupitzer2015,Mens2016,Tan2020}.
Finally, we tailored the definitions toward reconfiguration in robotic systems.

\subsection{Configuration Space (RQ2)}
\looseness=-1
For answering what parts of the software of a robotic system can be reconfigured and at what granularity (RQ2), it is essential to identify what are the reconfigurable elements in the robotic frameworks, how these are specified, and what dependencies might exist~\citep{Mens2003,Eddin2013}.
The \textit{Code Base} refers to the structuring of these reconfigurable elements of the robotic system.

To allow consistent and domain-independent investigation of the supported granularity, in this work, we use the following \edit{three} levels of granularity that are oriented on the naming from a prior study investigating variability in software ecosystem platforms~\citep{berger.ea:2014:ecosystems} and the observations from the SLR:
(i) A \textit{Composite Unit} represents a larger software entity, which can function independently of other units and usually aggregates and controls further smaller units,
(ii) reconfiguration takes place on the granularity of a single class or function that usually cannot function independently (\textit{Basic Unit}), e.g., using a different perception mechanism provided by the same composite unit, and
(iii) the value of a configuration attribute of a \textit{Unit parameter}, e.g., a parameter of a method or an environment variable, is changed.
\emph{Manifests} provide information about these reconfigurable entities such as their identifier or parameters required at reconfiguration, e.g., to identify a composite unit that is suitable for a specific task.

\subsection
{Configuration Space Model (RQ2)}
For safe reconfiguration, it is essential to know the space in which a reconfiguration of the code base can take place and what are valid configurations~\citep{Svahnberg2005SPE}.
To this end, these reconfigurable elements are represented using the notation of \textit{features}~\citep{kang1990feature} that serve as an abstraction of the \textit{code base}, therefore, providing a conceptual view on what elements of a robotic system can be reconfigured (RQ2).
Thereby, multiple \textit{features} can have dependencies among each others, e.g., one feature requiring or excluding another feature, which can limit what can be reconfigured.
Therefore, besides the reconfigurable elements, one has to know all possible combinations of these~\citep{Mens2016}.
\textit{Features} and their \textit{dependencies} are usually specified in a \emph{Configuration space model}~\citep{berger2013study}.
We are particularly interested in the languages supported by the frameworks for specifying \emph{Configuration space models}, as they are the main artifact used by developers for specifying the configuration space and could be analyzed by tools.

\subsection
{Encapsulation (RQ2)}
For reconfiguration in terms of turning features on or off, reconfigurable assets must be well encapsulated and allow one to access reconfigurable elements.
	%
	To this end, different \emph{Interface mechanisms}~\citep{berger.ea:2014:ecosystems,Tan2020} can be used to interact with reconfigurable elements.
These can range from well-defined APIs to system-wide messaging services.
%
Besides knowing which mechanisms the different robotic frameworks provide to allow communication among the reconfigurable elements, we also have to know in which way the interfaces of reconfigurable elements are specified so that developers can use the \emph{Interface specification}.

\subsection
{Interactions (RQ3)}
To understand reconfiguration, particularly, in terms of the mechanisms that can be used for implementing reconfiguration (RQ3), the management of interactions is essential.
To allow interaction at runtime, the statically defined interface usages have to be bound to the concrete running instances and have to be updated at each reconfiguration.
For executing the bindings, a reconfigurable system needs a \emph{Run-time Manager} that performs this binding~\citep{berger.ea:2014:ecosystems}.
Consequently, we are interested in how the robotic frameworks implement interaction binding. 

While the already considered interface mechanisms are a static view on specifying interaction, we also need to know the \emph{Interaction mechanism} using which dynamic interaction with reconfigurable assets is realized.
Interactions among basic units require \emph{Interaction binding} for identifying and binding the concrete target, which can happen at different times ranging from static binding to dynamic binding~\citep{Fritsch2008}.

\subsection
{Trigger \& Logic (RQ3)}
\label{sec:trigger}
Reconfiguration consists of three essential steps: planning, decision, and execution \citep{Tan2020}.
These steps must be implemented using appropriate mechanisms (RQ3), which, in the best case, are provided by robotics frameworks.

A reconfiguration trigger is a set of conditions that, if true, activate one or more reconfiguration actions \citep{Soria2009}.
We need to know what logic can be used to specify reconfiguration triggers \citep{Mens2003,Krupitzer2015} and how the reconfiguration will be executed \citep{Mens2003}.
We are interested in what support the robotics frameworks provide to developers for specifying \emph{Decision Criteria} to define when and how to reconfigure a robotic system.

As also confirmed by our SLR, decision criteria can be defined in various ways~\citep{Krupitzer2015}:
(i) model-based specification of the reconfiguration, e.g., a statemachine,
(ii) rule-based triggering of reconfiguration,
(iii) goal-based reconfiguration, and
(iv) frameworks, which offer utility functions that simplify the implementation of reconfiguration.

The reconfiguration itself can then be realized in different ways.
In the literature~\citep{Mens2003,Fritsch2008,Krupitzer2015}, we can find \edit{three} different \emph{Change types} of reconfiguration at runtime that correspond to a robots environment and the three software granularities in the asset base:
(i) changing parameter values to permanently influence the internal control flow of units (\textit{parameter}),
(ii) changing a functionality while keeping its interface, and
(iii) structural reconfiguration (\textit{structure}), which changes the running system's structure, e.g., by loading or unloading a composite unit.
\smallskip

This conceptual model allows us to systematically capture how reconfigurable elements are specified, how these interact with each other, how the reconfiguration logic can be specified, and which reconfiguration support robotic frameworks provide.

\section{Motivations for Reconfiguration (RQ1)}
\label{sec:rq1}
First, we want to get a better understanding of what scientists see as the reasons that can be addressed by reconfiguring robotic systems.
We assume that understanding their motivations, allows us to reason better on why they address reconfiguration as they did.
To this end, we only investigated the academic literature on reconfiguration to answer RQ1.

\subsection{Reasons for Reconfiguration in the Academic Literature}
\begin{figure}
	\centering
	\includegraphics[width=1.0\columnwidth]{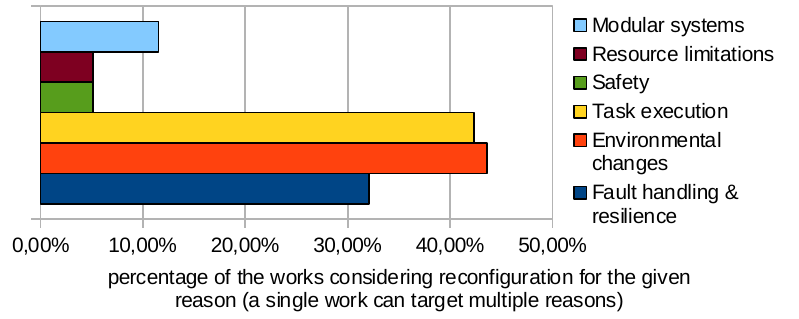}
	\caption{Reasons for reconfiguring a robotic system}
	\label{fig:slr:reasons}
\end{figure}

\noindent
\edit{
	Of the 78 papers, 27 papers (34.62\%) mention more than just one reason for reconfiguring robotic systems.
	On average 1.45 reasons are mentioned per paper.
	\Cref{fig:slr:reasons} shows the most popular reasons and their relative frequencies among all mentioned reasons.
	Only P4 and P67 mention no reason for reconfiguration.
}

\begin{description}
	\item[{Environmental changes \& task execution:}]
	As expected, reconfiguration mainly takes place to allow operation in dynamic environments \edit{(43.59\%)} and executing tasks \edit{(42.31\%)}.
	There seems to be a significant correlation between these two reasons, since \edit{15 papers}, which is by far the most common combination in our sample, mention both of them at the same time \edit{{\scriptsize (P1,
			P3,
			P17,
			P22,
			P30,
			P33,
			P41,
			P51,
			P56,
			P59,
			P68--69,
			P73--74, and P77)}}.
	Reconfiguration is needed (i) between tasks that require different hardware and software configurations of a robot and (ii) in a single task execution to successfully fulfill the task.
	An additional \edit{19 papers} need reconfiguration to react to environmental changes \edit{{\scriptsize (P12,
			P14,
			P20--21,
			P25,
			P29,
			P31,
			P36--37,
			P39,
			P52--53,
			P57,
			P62--65,
			P72, and
			P78)}} and \edit{16 papers} for changes as part of task execution \edit{{\scriptsize (P2,
			P5,
			P10,
			P15--16,
			P18,
			P24,
			P34--35,
			P38,
			P40,
			P43,
			P48,
			P50,
			P66, and
			P70)}}.

	\item[{Fault handling \& resilience:}]
	Reacting to events such as a sensor returning constantly faulty data, is the third most frequently mentioned reason for reconfiguration (32.05\%).
	In this case, the aim is usually to reconfigure the system so that it does not use the faulty sensor any longer by replacing it with a spare sensor, i.e., in software switching to a different sensor
	\edit{{\scriptsize (P1,
			P6--7,
			P11,
			P13,
			P26--28,
			P42,
			P46--47,
			P49,
			P54,
			P57--62,
			P64,
			P69--70,
			P75--77)}.}

	\item[{Modular systems:}]
	More than 11\% of all papers claim to need reconfiguration for developing modular systems \edit{{\scriptsize (P3,
			P13,
			P19,
			P23,
			P30,
			P32--33,
			P43, and P70)}.}
	To this end, they consider modularity at development time and hardware modularity at runtime, e.g., a robot that can change the actuators it uses, which triggers the reconfiguration of the running software.
	While we consider the second also as reconfiguration, in our understanding, building a software system at development time from different modules is not reconfiguration, but general design-time variability~\citep{chen2011systematic,Apel:2013wg,berger.ea:2020:emse}.
	Related to this, papers \edit{P23 and P70} mention, besides other reasons, also maintenance or updates at runtime as a reason for reconfiguration, e.g., a human having to manually perform ad-hoc reconfigurations to allow a robot to complete its task.

	\item[{Resource limitations:}]
	As one would expect due to the ever-increasing computational capabilities, it seems that limited resources, often considered as the main driver for reconfiguration, have become a much less relevant challenge in the last decade.
	Still, a significant number of papers mention limited resources as a reason for reconfiguring a robotic system, but it is mainly the older papers considered in our SLR that mention this reason.
	First, in the two oldest papers \edit{{\scriptsize (P8 and P9)}}, which are all from 1999, it is due to generally limited computing capacities. Later, in \edit{P24 from 2006 and P71 from 2012} the resource limitation is due to more sophisticated tasks, such as speech-based interaction with robots.
	The most recent paper of these \edit{{\scriptsize (P47)}}, discusses limited resources in terms of network capacity for teams of interacting robots.

	\item[{Safety:}]
	Finally, responding to safety issues is mentioned as a reason for reconfiguration by \edit{four papers {\scriptsize (P44--45, P55, and P71)}}, which notably is the only mentioned reason in all \edit{papers except P71}.
\end{description}

\subsection{Discussion of Reasons for Reconfiguration}
In summary, the main driver for reconfiguration in the literature is robots performing complex tasks in diverse environments, which includes both reconfiguration to respond to environmental changes during task performance, as well as adapting the configuration of the robot according to task needs \edit{(64.1\% mention one of these two reasons)}.
Orthogonal to this, reconfiguration is also needed to ensure proper operation by enabling fault handling and increasing the robustness (which is partly closely related to reconfiguration due to environmental changes) of the robots.
\section{Reconfigurable Elements of Robotic Systems (RQ2)}
\label{sec:rq2}
We answer RQ2 both from the academic perspective, as represented by the academic literature, and from the perspective of robotics frameworks.
We first present our analysis of the academic literature with respect to RQ2, then our analysis of robotics frameworks, and finally discuss the observations of both together.

\subsection{Academic Literature on What Parts of a Robotic System Can Be Reconfigured}
Based on the discussions above, we identified three aspects relevant for discussing what software parts of a robotic system can be reconfigured.

\textit{1) Granularity at which a robot is reconfigured:}
Since reconfiguration can be thought of in completely different dimensions, ranging from changing the hardware a robot uses to small differences in how it will behave, we want to assess what granularity of reconfiguration is considered in the academic literature on automated reconfiguration of robotics software at runtime.
Again, we expect to extract a baseline that can be used in a comparison with reconfiguration of robots in practice.


\textit{2) How long is the intended lifespan a configuration before reconfiguring the robotic system again:} We assume that robotic systems are not only configured once but are reconfigured more or less frequently according to relevant circumstances. Therefore, we aim at assessing how often reconfiguration is expected to take place and how long a robotic system will stay in a configuration before it is reconfigured again.

We classified the investigated papers according to these aspects to capture the academic perspective on RQ2.
In what follows, we present these classifications.

\subsubsection{Granularity of Reconfiguration}
\noindent
The investigated papers use many different terms for naming reconfigurable entities of robotic systems.
On one side, there are domain-specific terms, such as \textit{ROS node} due to the considered robotic framework.
On the other side, commonly used names, such as \textit{component}, are used differently among the papers.
This resulted either in different terms being used for referring to the same granularity or one term referring to different granularities in multiple papers.
To compare granularities across multiple papers, we assigned them to four categories based on the semantics described in each paper.
We oriented the definition of these categories on definitions from the literature~\citep{berger.ea:2014:ecosystems}.


\begin{figure}
	\centering
	\includegraphics[width=\columnwidth]{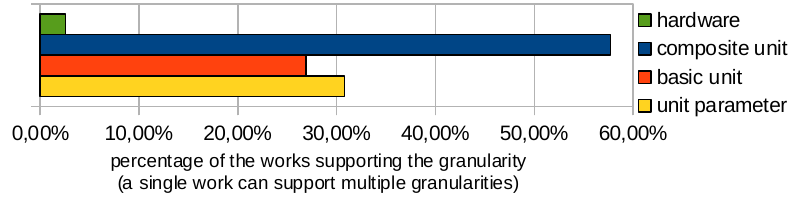}
	\caption{Granularity at which a reconfiguration is supported by the techniques presented in the literature}
	\label{fig:slr:granularity}
\end{figure}

\looseness=-1
\Cref{fig:slr:granularity} shows what percentage of the considered papers supports which level of granularity.
Thereby, a single paper can support multiple levels.
However, with an average of \edit{1.18 levels} per paper, the papers that consider multiple levels \edit{{\scriptsize (P3, P36, P41, P51, P58, P60--62, P67--68, P73, and P78)} are a minority.}
Overall, the literature seems to focus on reconfiguration at a coarse-grained level.

\begin{description}

	\item[Composite unit:] More than half of the papers \edit{(57.69\%)}, support reconfiguration \edit{of} larger software entities, which can function independently of other units, such as loading or unloading entire components of the system \edit{{\scriptsize (P1--4,
			P7,
			P13--22,
			P25,
			P27,
			P31,
			P33,
			P36--38,
			P40--42,
			P44,
			P47--48,
			P51--53,
			P57,
			P61--67,
			P69--71,
			P75, and P77--78)}.}

	\item[Basic unit:] \looseness=-1\edit{In more than a fourth of the papers (26,92\%),}
	reconfiguration takes place on the granularity of a single class or function, e.g., using a different perception mechanism provided by the same composite unit \edit{{\scriptsize (P5,
			P8--9,
			P12,
			P23,
			P26,
			P49--51,
			P54--56,
			P58--62,
			P68, and
			P72--74)}.}

	\item[Unit parameter:] The fine-grained reconfiguration of concrete parameters, such as changing the value of a field that represents a state, is considered by \edit{30.77\% of the papers {\scriptsize (P3,
			P6,
			P10--11,
			P24,
			P28--30,
			P32,
			P34--35,
			P39,
			P43,
			P45--46,
			P51,
			P58,
			P60,
			P62,
			P67--68,
			P73,
			P76, and
			P78)}.}
\end{description}

\subsubsection{Lifespan of Configurations}

\begin{figure}[b]
	\centering
	\includegraphics[width=\columnwidth]{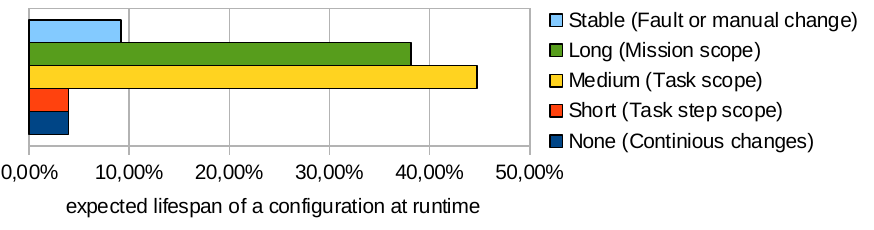}
	\caption{Lifespan of configurations}
	\label{fig:slr:lifespan}
\end{figure}

In the literature, configurations are typically meant to be alive for a relatively long time. Still, reconfiguration is considered to happen rather often and is not seen only in exceptional cases. We identified six different lifespans of configurations, whose popularity is shown in \cref{fig:slr:lifespan}.

\begin{description}

	\item[Stable:] The configurations are stable for a very long time, and reconfiguration is only performed due to exceptional events at runtime.
	These events are usually errors, such as sensors returning implausible values when hardware fails \edit{{\scriptsize (P26, P28, P47)}}.
	Also included in this category are reconfigurations that require a reboot of the robot \edit{{\scriptsize (P4)}} or a manual trigger by the user \edit{{\scriptsize (P4, P19, P32, and P47)}}, for example by manually attaching or detaching some hardware. Further cases comprise the generation of
	reconfigurations that require the robot to be placed in a non-dynamic state, such as stopping movements before executing a reconfiguration \edit{{\scriptsize (P4)}.}

	\item[Long:] The configuration is changed in the case of major environmental changes that are likely to occur, but do not occur frequently \edit{{\scriptsize (P21, P25, P37, P63--65, and P78)}}, or when a new configuration is required to perform a mission that may include multiple tasks \edit{{\scriptsize (P2, P15, P16, P18, P24, P43, P18, and P66)}}.
	\edit{Five papers {\scriptsize (P17, P22, P30, P68, and P74)}} consider major environmental changes as well as new mission requirements.
	In summary, we assume that a configuration will be used for the duration of an entire mission.
	Failures that are likely to occur in practice, but not frequently, may also lead to reconfigurations whose target configuration will be alive for a long time \edit{{\scriptsize (P11, P13, P42, P46, P49, P58, P64, and P75)}}.
	Similarly, this lifespan also applies to some types of resource constraints \edit{{\scriptsize (P8 and P24)}} and runtime maintenance~\edit{{\scriptsize (P23)}}.

	\item[Medium:] Already smaller environmental changes \edit{{\scriptsize (P12, P14, P20, P29, P31, P39, P52--53, P57, and P72)}} are likely to be addressed by reconfiguration, or reconfiguration is needed within a mission to execute the tasks of which the mission consists \edit{{\scriptsize (P5, P10, P35, P38, P40, P50, and P70)}}.
	Both of these two variants are considered at the same time by eleven papers \edit{{\scriptsize (P1, P3, P33, P41, P51, P56--57, P59, P91, P69, and P77)}.}
	\edit{Three papers {\scriptsize (P7, P27, and P54)}} consider faults that are likely to occur more often;
	and \edit{six papers {\scriptsize (P57, P59, P61, P69--70, and P77)}} consider them in addition to environmental changes or mission requirements. Also, resource limitations {\scriptsize (P9)} or safety reasons \edit{{\scriptsize (P44--45, and P55)}} are considered to trigger a reconfiguration that results in a configuration that is assumed to be alive for a medium time. Accordingly, we define the scope of a configuration with a medium lifespan to be of the lifespan of a single task.

	\item[Short:] To execute a single task of a mission already multiple reconfigurations might be needed. Configurations have a relatively short lifespan and are reconfigured quite frequently.
	\edit{P60} reconfigures for each step of an executed task; in \edit{P71} every minor change will be addressed by reconfiguring the robot; and in \edit{P73} a model used in a MAPE-K feedback loop is frequently updated.

	\item[None:]The configuration is constantly changing, and therefore, has no lifespan.
	In most cases, this is realized by continuously updating parameters \edit{{\scriptsize (P6, P34, and P76)}}.
	In \edit{P6}, feedback in the form of parameter reconfiguration is added to a linear time-invariant system model;
	in \edit{P34} the parameters of a time-division multiple access protocol are reconfigured; and in \edit{P76} the parameters of a force allocation matrix are reconfigured, which is used to define how the thruster configuration affects the dynamics of the UX-1 robot, an under water robot for exploration of flooded mine tunnels~\citep{Fernandez2019}.
\end{description}

\noindent
\looseness=-1
In summary, reconfiguration in the literature is considered to be executed quite frequently, but not continuously.
Configurations are mostly considered to be alive for a long time, having an entire mission as scope \edit{(38.16\%);}
or for a medium time with an entire task of a mission as scope \edit{(44.74\%).}
Considering the most frequent granularity, which is the reconfiguration of composite units, the time needed to execute a reconfiguration could be a major reason that stands against more frequent reconfiguration.
This reason was also mentioned in some of the papers, among others in detail in \edit{P4}.

\subsection{Reconfigurable Elements in Robotics Frameworks}
\label{sec:rq2:frameworks}
We investigated the robotics frameworks to determine which elements of a robotics system they support in terms of reconfiguration.
We based this analysis on our conceptual model introduced in \cref{sec:model}.
\Cref{tab:mechanisms} shows a summary of the four major robotic frameworks we analysed according to the introduced conceptual model.

\begin{table*}
\caption{Reconfiguration elements in robotic frameworks\label{tab:mechanisms}}
\scriptsize
\setlength{\tabcolsep}{2pt}
\begin{tabularx}{\linewidth}{
  l@{\hspace{1mm}}
  >{\hspace{1mm}}l
  >{\raggedright\arraybackslash}X
  >{\raggedright\arraybackslash}X
  >{\raggedright\arraybackslash}X
  >{\raggedright\arraybackslash}X
}
&  & \textbf{\textsf{ROS/ROS2}} & \textbf{\textsf{OROCOS}} & \textbf{\textsf{YARP}} & \textbf{\textsf{RobMoSys (SmartSoft)}} \\

\toprule

\multirow[c]{14}{*}{\rotatebox{90}{\textbf{Configuration Space}}}%
& \textbf{Asset Base}\smallskip & & & & \\
& \hspace{8pt}Basic units\smallskip
	& dynamic libraries (ROS plugin/nodelet)
	& dynamic libraries (plugin services)
	& dynamic libraries (YARP plugin services), static library (YARP module)
	& service
	\\
& \hspace{8pt}Composite units\smallskip
	& executable programs\newline (ROS node)
	& dynamic libraries (component)
	& c++ executable programs (component)
	& component
	\\
& \hspace{8pt}Unit parameters
	& ROS parameter
	& data flow port
	& YARP port
	& component parameter
	\\\cmidrule{2-6}
& \textbf{Configuration model} & & & & \\
& \hspace{8pt}Features\smallskip
	& node, plugin/nodelet
	& component, plugin
	& component, plugin
	& component
\\
& \hspace{8pt}Language & N/A & N/A & N/A & Variability Modeling Language (VML)
\\\cmidrule{2-6}
& \textbf{Manifest (Schema)}
	& XML-based DSL (launch file, plugin description file)
	& XML deployment file
	& XML-based DSL (YARP manager.ini, plugin manifest)
	& component model,\newline system configuration model
\\
\midrule

\multirow[c]{9}{*}{\rotatebox{90}{\textbf{{Encapsulation}}}}
& \textbf{Interface mechanism}
	& IDL-based messages (node), base class API (nodelet/plugin),\newline ROS parameters 
	& TaskContext API, IDL-based component services 
	& IDL-based messages (programs), base class API (RFModule) 
	& provided/requires services 
\\\cmidrule{2-6}
& \textbf{Interface specification}
	& Documented interfaces (message types and package descriptions) in public repositories
	& Documented interfaces in the Orocos Component Library
	& Documented interfaces
	& SmartMARS Metamodel (communication objects + communication patterns)
\\

\midrule
\multirow[c]{8}{*}[-1mm]{\rotatebox{90}{\textbf{{Interactions}}}}%
& \textbf{Run-time Manager}
	& ROS master, DDS
	& OROCOS DeploymentComponent
	& YARPserver, YARPmanager
	& SmartEventServer, SmartParameterMaster
\\\cmidrule{2-6}
&\textbf{Characteristics} &&&&\\

& \hspace{8pt}Mechanism\smallskip
	& publish/subscribe, client/server, parameters
	& data-flow,\newline client/server
	& data-flow,\newline client/server
	& publish/subscribe,\newline client/server\\

& \hspace{8pt}Binding mode\smallskip
	& dynamic & dynamic & dynamic & dynamic\\

\midrule
\multirow[c]{17}{*}{\rotatebox{90}{\textbf{{Trigger \& Logic}}}}
& \textbf{Decision Criteria}\smallskip 	& & & & \\
& \hspace{8pt}Models\smallskip
	& state machine (ROS SMACH)\newline behavior trees (py\_tress\_ros \& BehaviorTree.CPP)
	& state machine
	& behavior trees
	& dynamic statecharts
\\
& \hspace{8pt}Rules/Policies\smallskip 	& N/A & N/A & N/A & VML: ECA rules\\
& \hspace{8pt}Goals\smallskip 			& N/A & N/A & N/A & SmartTCL (Task Coordination Language)\\
& \hspace{8pt}Utility function
	& ROS APIs
	& RTT:APIs
	& YARP APIs
	& N/A
\\\cmidrule{2-6}
& \textbf{Change Type}\smallskip 	&  & & & \\
 & \hspace{8pt}Parameter\smallskip
 	& dynamic$\_$reconfigure
 	& RTT::TaskContext (data flow port)
 	& YARP::os::BufferedPort
 	& SmartParameterMaster/\newline Client (parameter)
 \\
 & \hspace{8pt}Functionality\smallskip
 & pluginlib (plugin)
 & N/A
 & N/A
 & N/A
 \\
 & \hspace{8pt}Structure\smallskip
 	& roslaunch (node), pluginlib + nodelet (plugin/nodelet)
 	& RTT::Scripting (plugin)
 	& YARP::dev::DriverCreator
 	& SmartTask (component)
 \\

\bottomrule

\end{tabularx}
\end{table*}



\subsubsection
{Configuration Space}
All frameworks provide structures for specifying reconfigurable assets and realize the three different element kinds that can be part of the \emph{Code Base}.
Most frameworks provide multiple realizations of \textit{basic units}.
ROS, OROCOS, and YARP realize \textit{basic units} as shared or dynamically loadable libraries.
With \emph{plugins} and \emph{nodelets}, ROS implements two kinds of composite units.
Plugins allow us to load additional functionalities into the executed methods.
In contrast to this, \textit{nodelets} can be executed in separate threads in parallel to the current execution.
In RobMoSys, \textit{basic units} are the provided services, which are specified as models according to a metamodel of RobMoSys.
Depending on the concrete implementation, e.g., SmartSoft, the code for a service can be generated and loaded dynamically.

The \textit{composite units} are realized in ROS and YARP as executable programs that use the frameworks for inter-unit communication and for accessing basic units.
In OROCOS, composite units provide the basic infrastructure to make a system out of pieces of code that can interact via data and events.
RobMoSys services are gathered in components. 

\textit{Unit parameters} are realized differently across the frameworks.
In ROS, composite units can have parameters that are managed globally and must be actively accessed by the composite units.
OROCOS and YARP are based on input and output ports of composite units that are explicitly connected with each other, data directly flows between composite units according to their linking.
In RobMoSys, components have a model-based parameter specification.

\subsubsection
{Configuration Space Model}
In all frameworks, the composite units realize features that can be turned on or off as they are considered in the literature~\citep{kang1990feature}.
In ROS, OROCOS, and YARP, even the basic units are features.

All frameworks work with XML-based \emph{Manifests (Schema)} for specifying the reconfigurable elements (\emph{features}).
Thereby, the \emph{Manifests} can be distributed over multiple files, e.g., one per composite unit.
However, only RobMoSys provides a detailed metamodel that defines the syntax of the \emph{Manifest (Schema)}.
Similarly, among the robotic frameworks, only RobMoSys provides an explicit model of the configuration space, despite its importance for safe reconfiguration.
This model can be specified using the \textit{Variability Modeling Language} (VML)~\citep{Schlegel2013}.
For ROS, there are at least third-party extensions that provide such capabilities, such as HyperFlex~\citep{SLR-2016-Brugali-1}.

\subsubsection
{Encapsulation}
All robotic frameworks provide suitable interface mechanisms to realize well-en\-capsulated reconfigurable assets.
%
Except for RobMoSys, the only explicitly specified \emph{Interface mechanism} is an abstract class that must be implemented by the reconfigurable assets.
RobMoSys provides a metamodel for describing communication objects and communication patterns, e.g., publish/subscribe or client/server.
However, while the interfaces are clearly defined, this does not necessarily imply that their semantic usage is equally simple.
For example, interfacing with the ROS navigation stack can require non-trivial sequences of messages.

Except for RobMoSys, all frameworks rely on manual documentation of the interfaces.
ROS and OROCOS encourage developers to provide such \emph{Interface specification} by making it an essential part of public ROS repositories, and of the OROCOS component library.
In practice, however, the documentation of packages is often outdated and not maintained.
It is well-known that good documentation is needed for effectively working with software \citep{Aghajani2019}, and, therefore, this lack in up-to-date documentation is a significant challenge for the implementation of robotic systems.

	\subsection{Discussion of Reconfigurable Elements}
	Three distinct levels of granularity were identified at which robotic systems can be reconfigured.
	The first level comprises concrete configuration parameters that can be assigned values.
	The second level encompasses basic units that provide concrete functionalities but are not independently executed.
	The third level encompasses composite units that provide larger software entities that can comprise multiple basic units and are executable on their own.
	While the majority of literature focuses on the structural reconfiguration of robotic systems, primarily at the granularity of composite units, robotic frameworks offer only limited support for this granularity concerning reconfiguration.
	Technically, the frameworks facilitate the structuring of robotic systems according to different relevant granularities for reconfiguration.
	However, they lack the means to explicitly specify configurable elements and possible constraints on them.
	This represents a significant challenge to their reconfiguration since possible interactions are not explicitly captured and, therefore, difficult to handle.

\section{Reconfiguration Mechanisms (RQ3)}
\label{sec:rq3}
To answer RQ3, we conducted an investigation of all three kinds of artifacts. First, we present the investigation of the academic literature that represents the state-of-the-art concerning RQ3. Then, we present the corresponding state-of-practice, in terms of robotic frameworks and robotic (sub-)systems. Finally, we answer RQ3 concerning all three artifacts.

\subsection{Academic Literature on Reconfiguration Mechanisms}
\label{sec:rq3:slr}
We reviewed the academic literature on reconfiguration for the following three aspects of reconfiguration mechanisms needed to develop reconfigurable robotic systems.

\textit{1) Specification of the reconfiguration logic:} Reconfiguration is a multi step process, before actually executing a reconfiguration, the robotic system must identify the need to do so and plan on how to react. The latter two steps must be implemented in some way in a robotic system and we are interested in identifying the techniques that have been proposed in the scientific literature.

\textit{2) Ensuring the validity of reconfigurations:} To allow safe and secure operation of robotic systems, it is essential that they are always in a valid configuration. Invalid configurations can cause security issues~\citep{Peldszus2018} as well as safety issues.
	Unfortunately, it has been shown that already statically ensuring the validity of configurations is challenging~\citep{Yin2011}.
	Therefore, we are interested how this issue can be addressed in the robotics domain.

\textit{3) Cost of reconfiguration:} The different mechanisms for specifying, validating, and executing reconfigurations usually come with some cost for executing these tasks. However, in what form these cost arise, what dimensions of cost must be considered for reconfiguration, and their impact are unclear so far.

\subsubsection{Specification of Reconfiguration}
\label{sec:rq3:slr:specification}

\begin{figure}[t]
	\centering
	\includegraphics[width=\columnwidth]{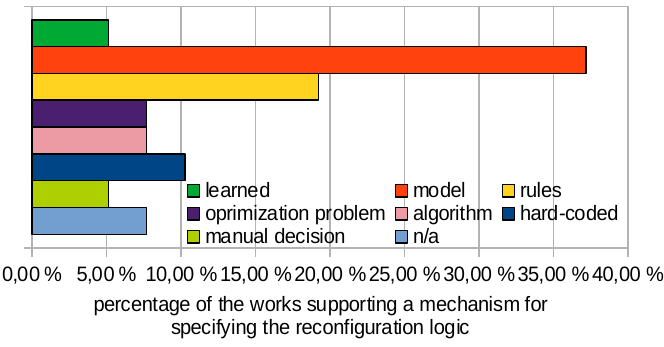}
	\caption{Specification of reconfiguration logic}
	\label{fig:slr:specification}
\end{figure}

\noindent
In the investigated papers, we found seven different approaches to specifying the reconfiguration logic of a robotic system.
\Cref{fig:slr:specification} shows the approaches and how often these were implemented in the investigated papers.
Only \edit{6 papers} do not explicitly consider the specification of the reconfiguration logic \edit{{\scriptsize (P33,
	P43,
	P51,
	P63--64, and
	P75)}}, which is indicated by \textit{n/a}.

\begin{description}
	\item[Learning:] In \edit{4 papers}, no explicit specification of the reconfiguration logic is needed, since the papers focus on automatically learning when and how to reconfigure the robot \edit{{\scriptsize (P41--42,
		P52, and
		P72)}}.

	\item[Model:] The reconfiguration space and task or mission requirements are explicitly modeled.
	There are techniques that rely on two connected models. First, a model that captures the possible configuration space, i.e., alternative configurations for various situations are explicitly defined or the configuration space is specified using feature models \citep{kang1990feature}.
	Second, a model of the tasks or missions to be executed that is connected to needed elements from the latter model via explicit references.
	Some central unit selects the best configuration based on the current task and situation \edit{{\scriptsize (P2,
		P5,
		P8--10,
		P16--18,
		P20--22,
		P25,
		P27--29,
		P39,
		P50,
		P53,
		P56,
		P59,
		P61,
		P66,
		P70,
		P73,
		P76--78)}.}
	\edit{P38} express the configuration logic as executable models, such as a state machine, avoiding the need for some central unit for selecting target configurations; and \edit{P49} uses a mathematical model.


	\item[Rule:] The reconfiguration is explicitly specified in rules, whose application conditions are monitored at runtime and the respective rule is executed as soon as the condition applies \edit{{\scriptsize (P15,
		P24,
		P31,
		P35--36,
		P40,
		P48,
		P54,
		P57--58,
		P60,
		P68--69,
		P71, and
		P74)}}.
	The rule then changes to a specific configuration; or a specific reconfiguration action will be executed, such as replacing a composite unit.
	Rules are typically encapsulated in separate artifacts that can be maintained without touching the implementation of the robotic system.
	In addition, domain-specific languages are often used to facilitate the specification of these rules.


	\item[Optimization problem:] \edit{Six papers} treat the reconfiguration as an optimization problem of the kind of finding the best configuration that fulfills some given criteria \edit{{\scriptsize (P1,
		P26,
		P30,
		P46--47, and
		P55)}.}
	For example, in \edit{P26} the developers define a multi-dimensional constraint problem for which an optimal solution in terms of a target configuration must be identified.
	In contrast to the papers that are based on models of the robotic system, which are likely to select a new configuration also based on the outcome of some optimization problem, in these papers, the optimization problem is hard coded and is not based on the interpretation of a model. 

	\item[Algorithm:] In another \edit{six papers {\scriptsize (P6--7,
		P11--12,
		P34, and
		P45)},} the reconfiguration logic is realized as an algorithm determining the conditions that trigger a reconfiguration.
	The algorithms considered in this category are typically fixed reconfiguration algorithms for specific reconfiguration tasks, e.g., error detection algorithms to detect a faulty component and reconfigure the system (usually by disabling the faulty component and enabling a backup).
	These algorithms cannot be tailored to project-specific needs by developers but are intended to be used as they are for a specific reconfiguration task.

	\item[Hard-coded logic:] In \edit{eight papers {\scriptsize (P13--14,
		P19,
		P32,
		P37,
		P62,
		P65, and
		P67)},} the reconfiguration logic is a hard-coded part of the implementation of the robotic system and typically consists of if-then statements.
	Unlike fixed algorithms, the logic is tailored towards the specific robotic system and is maintained as part of the source code of the robotic system.

	\item[Manual selection of configuration:] Finally, \edit{four papers {\scriptsize (P3--4,
		P23, and
		P44)}} do not consider the specification of when and how to reconfigure.
		\edit{
			These papers focus \edit{on} automatically reconfiguring into a target configuration.
			For example, manual triggers, such as attaching a tool (e.g., a screwdriver) to the robot serve as a trigger to perform a automated reconfiguration into a  specific configuration (e.g., loading the software modules needed to work with the screwdriver).
			While the target configuration is given manually, the system is configured automatically, among others involving activities such as bringing the system into a safe state before changing its configuration {\scriptsize(P4)}.
		}
\end{description}

\noindent
\looseness=-1
Altogether, basing the reconfiguration logic on models of the robotic systems is the most popular way of specifying the reconfiguration logic with \edit{37.18\% of the papers.}
The relatively similar, but much simpler specification of the reconfiguration logic in reconfiguration rules follows with \edit{19.23\%} of the papers.
\edit{The reconfiguration logic being hand-written in code is considered by 10.26\% of the papers.}
All other specifications of the reconfiguration logic account for only between \edit{5.13\% and 7.69\%} of the papers and could be considered less relevant or at least less popular for the development of dynamically reconfigurable robotic systems.

Both rule- and model-based reconfiguration are built on user-defined specifications of the logic itself or of the system.
For these papers, we found mostly \textit{Domain-Specific Languages (DSLs)}~\citep{Wasowski2023} for defining when and how to reconfigure.
These DSLs range from providing simple mappings between tasks and composite units to languages that support complicated conditions that are solved to determine a suitable configuration within the robot's configuration space.
Unfortunately, many papers only mention a type of specification that is not detailed in the paper.
Usually these specifications are mappings between goals or tasks and reconfigurable assets of the robot that are suitable for achieving the goal or performing the task.

\subsubsection{Validity of Reconfigurations}

\begin{figure}
	\centering
	\includegraphics[width=\columnwidth]{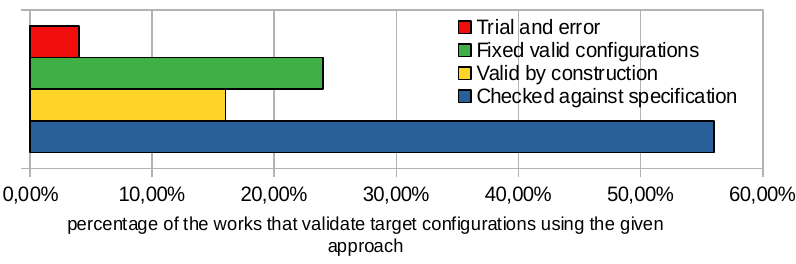}
	\caption{Assurance of target configuration validity}
	\label{fig:slr:validation}
\end{figure}

While all of the \edit{78 reviewed papers} describe approaches that enable reconfiguration, only \edit{28 of them} explicitly consider ensuring that a reconfiguration will result in a valid configuration of the robot.
\Cref{fig:slr:validation} shows the identified validation approaches and how often these appear in the investigated papers.

\begin{description}
	\item[Checking against specification:] \edit{In 14 papers {\scriptsize (P7,
		P45,
		P48,
		P51--52,
		P55,
		P61,
		P66,
		P69--70,
		P72--74, and P77)},} the validity of a reconfiguration is checked against a specification before executing the reconfiguration.
		\edit{In four of these papers (28.57\% of the cases)} this specification is a feature model~\citep{kang1990feature} \edit{{\scriptsize (P45, P48, P55, and P74)}}, \edit{in another four cases it is an architecture specification {\scriptsize (P52, P72, P74, and P77)}}, in \edit{three cases} the checking is performed against a state machine containing the reconfiguration logic \edit{{\scriptsize (P51, P70, and P73)}}, and in another \edit{three cases} general well-formedness rules are used \edit{{\scriptsize (P61, P66, and P69)}}.
	\edit{In P7}, the validity is only checked in regard to whether application-specific timing constraints are met, which differs from the other papers in this category.

	\item[Fixed valid configurations:] \edit{Six papers {\scriptsize (P46,
		P53,
		P63,
		P65,
		P76, and
		P78)}} use a fixed configuration space that only contains valid configurations.
	Either the reconfiguration algorithm can only chose from a predefined set of valid configurations \edit{{\scriptsize (P46, P53, P76, and P78)}}, or the reconfiguration space consists only of simple replacements that are always valid \edit{{\scriptsize (P63 and P65)}.}

	\item[Valid by construction:] \edit{Four papers {\scriptsize (P4,
		P56, and
		P59--60)}} ensure the validity of target configurations by the way they construct their reconfiguration mechanism.
	Compared to using a fixed number of valid configurations to choose from, in these papers, additional constraints to be fulfilled by possible target configurations must be considered, and the reconfiguration logic cannot just choose any valid configuration to fulfill the task.
	\edit{In P4,} the reconfiguration of a robotic system realized based on port-based objects, which are composite units that have defined numbers of input and output ports using which they can be connected with each other.
	Validity is achieved when all input ports of port-based objects are connected to output ports.
	\edit{In P59 and P60}, validity must be manually verified at development time, while in \edit{P56}, validity is encoded in the optimization problem to be solved at runtime, resulting in only proposing valid~reconfigurations.

	\item[Trial and error:] \edit{P12 proposes} a trial-and-error-based reconfiguration method.
	This method involves attempting various configurations until the robot has reconfigured into a configuration that allows the execution of an intended task.
\end{description}

\noindent
Overall, since below a third of all investigated papers consider checking the validity reconfigurations, validity seems to be a minor concern in the robotics community so far, although it is essential for the huge configuration spaces proposed in many papers.
Much work in this direction has already been done in the product line community, e.g., when focusing on finding optimal configurations~\citep{Henard2015,Guo2018, Pereira2021}.
Nevertheless, the more recent papers considered in our SLR already seem to integrate such results \edit{{\scriptsize (P45, P48, P51, P52, P55, P56, P70, P72--75, and P77)}}, the oldest of which \edit{{\scriptsize (P70)} dates from 2010.}

Besides this, \edit{nine} additional papers consider the validity of reconfiguration, but these are practically infeasible for more complex reconfiguration scenarios.
In particular, this comprises the papers that work with a fixed pool of valid configurations \edit{{\scriptsize (P46, P53, P63, P65, P76, and P78)}}, a constructive approach that limits the number of possible configurations \edit{{\scriptsize (P4)}}, or completely manual verification \edit{{\scriptsize (P59 and P60)}.}
However, these papers were mainly published between 1997 and 2006 and might represent a dated view on the validity of reconfigurations.
Still, \edit{four of these papers {\scriptsize (P46, P53, P76, and P78)}} were recently published between 2019 and 2022.

Nevertheless, we see a trend towards systematic validity checks against the design-time specification of the configuration space and task requirements, as \edit{56\% of the papers} dealing with target configuration validation (mainly the more recent ones) check the validity of reconfigurations against such specifications.

	\subsubsection{Cost of Reconfigurations}
	Roughly, one third of all investigated papers \edit{(36.62\%)} consider the cost related to reconfiguration in some form.
	\edit{In 51.85\%} of the papers, this cost arises from additional computations necessary for performing computations.
	All other papers do not explicitly mention a source of cost.
	The papers capture the cost in two different dimensions, were only \edit{P39} considers both dimensions:

	\begin{description}
		\item[Time:] With \edit{18 papers}, the majority considers the cost in terms of time needed for reconfiguration \edit{{\scriptsize (P9,
			P17,
			P38--39,
			P42,
			P44,
			P47--50,
			P52,
			P56,
			P61,
			P64,
			P66,
			P69--70, and P78)}.}
		Among others, taking too long for reconfiguration could lead to violating real-time constraints or having to interrupt the operation.
		\item[Resources:] \edit{Three papers} name the resource usage caused by a reconfiguration.
		\edit{P73} reports a static memory overhead due to running a reconfiguration engine.
		\edit{P27 and P39} generally mention that the usage of computing resources could impact the performance of a robotic system since they are partially blocked during reconfiguration.
	\end{description}

	\noindent
	Among the papers that consider the cost of reconfiguration, \edit{11 papers} do not further specify where these cost arise \edit{{\scriptsize (P9,
		P20,
		P22--23,
		P27,
		P38,
		P44,
		P50,
		P61, and
		P75--76)}.}
	The other papers identify two sources of cost:

	\begin{description}
		\looseness=-1
		\item[Planning:] The majority \edit{(13 papers)} consider the cost of planning of how to reconfigure the robotic system \edit{{\scriptsize (P17,
			P39,
			P46--49,
			P52,
			P56,
			P63,
			P66,
			P69--70, and
			P73)}.} The cost of planning a reconfiguration depends on the configuration space and the number of possible reconfigurations as well as the way how possible configurations are computed.
		Calculating an optimal solution in a large configuration space might not scale \edit{{\scriptsize (P46, P49, P56, P63, P66, P69, and P73)}.}
		However, assuming a sufficiently small configuration space, an optimal solution is still feasible, since the authors measured for \edit{P52} on average 5.57s generating reconfigurations followed by another 5.94s for model checking them ro generate an optimal target configuration.

	\looseness=-1
		Faster planning can be achieved by accepting non optimal target configurations or restricting the reconfiguration problem\,\edit{{\scriptsize (P17, P47, P49, and P70)}} or incremental reconfiguration planning \edit{{\scriptsize (P39)}}.
		If the planning is designed well, the time for planning can be reduced to times below six seconds\,\edit{{\scriptsize (P70)} or even 0.66 seconds\,{\scriptsize (P47)}.}

		\item[Execution:] Only \edit{three papers} consider the costs of actually executing a planned reconfiguration \edit{{\scriptsize (P42, P64, and P78)}.}
		\edit{P42 and P78} actually measured the cost of executing a structural reconfiguration of composite units, with the latter measuring 0.47 seconds to change a configuration and the other 4.2 seconds. In summary, the cost of executing a reconfiguration is not negligible, but can be relatively low.
	\end{description}

\subsection{Reconfiguration Mechanisms of Robotics Frameworks}
\label{sec:rq3:frameworks}
\noindent
To capture the state-of-practice concerning RQ3, we investigated what reconfiguration mechanisms robotics frameworks provide.
Again, we based this analysis on our conceptual model introduced in \cref{sec:model}, and \cref{tab:mechanisms} also contains the classifications of the robotics frameworks concerning reconfiguration mechanisms.

\subsubsection{Interactions}
\noindent All frameworks provide means to implement runtime interactions, e.g., data exchanges or calling functionality, among reconfigurable elements that are orchestrated by one or more \emph{Runtime Manager}.
RobMoSys uses  Smart\-Event\-Server for publish/subscribe communication and SmartParameterMaster for parameter-based communication.
%
ROS has a centralized ROS master, while ROS2 uses a decentralized data distribution service (DDS).
In YARP, every component periodically retrieves incoming messages from central framework entities.

\looseness=-1
The robotic frameworks provide different \emph{Interaction mechanisms}, allowing reconfigurable assets to interact at runtime.
They mainly use message-based communication; e.g., ROS nodes can publish and subscribe to messages based on topics, and composite units can subscribe to others in RobMoSys.
Furthermore, ROS, YARP, and RobMoSys offer client/server communication.
In contrast, the communication in OROCOS is purely parameter based and the entire control flow is handled by the framework.
%
In all robotic frameworks, \emph{Interaction binding} 
takes place when the source code statements implementing interactions are executed (late dynamic).

\subsubsection{Trigger \& Logic}
%
Only RobMoSys provides a wide variety of possibilities for specifying \emph{Decision Criteria} in terms of reconfiguration triggers and reconfiguration logic.
ROS, OROCOS, and YARP provide some kind of behavior model that is not explicitly intended to specify reconfiguration.
Instead, reconfiguration is mainly implemented in the composite units using the framework's utility functions (equivalent to the hard-coded specification of reconfiguration triggers and logic that we found in the literature (see \cref{sec:rq3:slr:specification})).
Consequently, developers are obliged to consider and address all potential side-effects, such as the time required for startup or shutdown of a component, and any intermediate states that may arise from this. In this regard, the frameworks offer minimal support in addressing these challenges.

\edit{
Of the \emph{Change types} in our conceptual model, all frameworks support reconfiguration of the \textit{structure} of a robot's software, which is also the most commonly observed change type in the SLR, and \textit{parameter} reconfiguration. 
Only ROS supports the \emph{Change type} of \textit{functionality} reconfiguration through its \texttt{pluginlib}.
}

In ROS, parameter reconfiguration is implemented in the \textit{dynamic\_reconfigure} library.
While ROS does not provide explicit support for structural reconfiguration, launching and terminating composite units is implemented in the \textit{roslaunch} library, which can also used for structural reconfiguration at runtime.
	Loading basic units is implemented in \textit{pluginlib}/\textit{nodelet} depending on the concrete realization of the basic unit that should be instantiated.
OROCOS offers the API \textit{TaskContext} for accessing and changing parameters and the API \textit{Scripting} for structural reconfiguration in terms of loading plugins.
A component called \textit{DriverCreator} manages the loading of components in YARP.
In the SmartSoft implementation of RobMoSys, parameter reconfiguration is managed by a \textit{SmartPara\-meter\-Master} and \textit{Smart\-Parameter\-Client}, while components are organized by the service \textit{SmartTask}.
However, the intention of RobMoSys is to approach reconfiguration using model-based techniques, where reconfiguration is specified in models rather than implemented in code; models are then executed to realize the behavior and reconfiguration of the robotic system.
\smallskip

In conclusion, the specification of reconfiguration triggers and logic, as well as their execution in robotic systems, is where we see the biggest mismatch between what is proposed in the literature and what robotic frameworks provide.
With the exception of RobMoSys, no framework provides the means to explicitly specify the reconfiguration space, nor to specify concrete reconfigurations.
To some extent, behavioral models, e.g., using statecharts or behavior trees~\citep{colledanchise2018behavior}, could be used to specify reconfiguration.
However, their main purpose is to describe actual behaviors in terms of self-adaptation.
As part of this, reconfigurations could be triggered, for example, in the implementation of an action node in a behavior tree.
In this case, however, the actual reconfiguration must still be implemented using the reconfiguration-related APIs of the frameworks.
In the end, the intended method is to hard-code the reconfiguration mechanisms using the reconfiguration-related APIs of the frameworks, e.g., for launching and terminating composite units from code.
This pure focus on source-code APIs of the frameworks is somewhat contradictory to the significant focus on specifying reconfiguration using DSLs and reference architectures for reconfigurable robots observed in our SLR.
However, one can clearly see that RobMoSys originates from an academic research project, as it provides exactly such specification formats for decision criteria.
However, these libraries are not explicitly defined in any framework with a focus on reconfiguration, but are intended for launch-time configuration.

\subsection{Reconfiguration in ROS (Sub-)Systems}
\label{sec:applications}
\noindent
To investigate how reconfiguration is actually implemented in practice and to derive best practices for implementing reconfiguration, we investigated the source code of open-source robotics (sub-)systems.
To this end, we started the investigation with a list of 115 open-source ROS-based (sub-)systems of Malavolta et al.~\citep{Malavolta2021}.
One of these repositories was no longer publicly accessible.
We filtered the accessible repositories for all (sub-)systems that use one of the identified ROS libraries (see Sections \ref{sec:rq2:frameworks} and \ref{sec:rq3:frameworks}) suitable to implement reconfiguration.
Since the libraries have to be explicitly imported, this was easily done by name matching.
We kept all (sub-)systems whose source files contain one of the following library names: (i) \textit{roslaunch}, (ii) \textit{nodelet}, (iii) \textit{pluginlib}, and (iv) \textit{dynamic\_reconfigure}.
This resulted in 48 (sub-)systems that potentially contain implementations of reconfiguration.
\Cref{tab:applications} shows a list of the investigated \mbox{(sub-)sys}\-tems and which libraries they use.
In some cases, one of the keywords was used in comments of the source code, e.g., in a description of how to manually launch the (sub-)system using \textit{roslaunch}.
We provide all details on the (sub-)systems and our findings in our replication package \citep{replication}.

\begin{table*}
	\caption{Investigated robotic (sub-)systems, the ROS libraries they use (\checkmark used library, (\checkmark ) only text match in the source code, -- not used), and the implemented reconfiguration (\checkmark implemented reconfiguration, (\checkmark) partially implemented reconfiguration, -- not implemented)}
	\label{tab:applications}
	\center
	\setlength{\tabcolsep}{3.5pt}
	\begin{smaller}
	\begin{tabular}{rl|cccc|ccc}
		\toprule
			&									& \multicolumn{4}{c}{Used ROS Libraries} 								& \multicolumn{3}{c}{Reconfiguration of} \\
		ID	& (Sub-)System						& dynamic\_reconfigure 	& nodelet 		& pluginlib 	& roslaunch 	& Unit Parameters & Basic Units & Composite Units \\
		\midrule
		1	& ani								& --					& --			& -- 			& -- 			& --			& --	& -- \\
		2	& autorally							& \checkmark			& \checkmark	& \checkmark	& --			& \checkmark & \checkmark & -- \\
		3	& avoidance							& \checkmark			& \checkmark	& --			& --			& \checkmark & -- & -- \\
		4	& bebop\_autonomy					& \checkmark			& \checkmark	& \checkmark	& --			& \checkmark & -- & -- \\
		5	& capabilities						& --					& \checkmark	& --			& --			& -- & -- & -- \\
		6	& cob\_command\_tools				& \checkmark			& --			& --			& --			& -- & -- & -- \\
		7	& cob\_control 						& \checkmark			& --			& \checkmark	& --			& \checkmark & -- & -- \\
		8	& cob\_environment\_perception		& \checkmark			& \checkmark	& \checkmark	& --			& \checkmark & \checkmark & -- \\
		9	& cola2\_core						& \checkmark			& --			& --			& \checkmark	& \checkmark & \checkmark & -- \\
		10	& ed								& --					& --			& \checkmark	& --			& -- & -- & -- \\
		11	& elfin\_robot						& \checkmark			& --			& \checkmark	& --			& \checkmark & -- & -- \\
		12	& evapi\_ros						& \checkmark			& --			& --			& --			& \checkmark & -- & -- \\
		13	& exotica							& --					& \checkmark	& \checkmark	& --			& -- & -- & -- \\
		14	& flexbe\_behavior\_engine			& --					& --			& --			& \checkmark	& -- & -- & -- \\
		15	& franka\_ros						& --					& --			& \checkmark	& --			& -- & -- & -- \\
		16	& free\_gait						& --					& --			& \checkmark	& \checkmark	& -- & \checkmark & -- \\
		17	& FusionAD							& \checkmark			& \checkmark	& \checkmark	& --			& \checkmark & -- & -- \\
		18	& grvc-ual							& --					& --			& --			& (\checkmark )	& -- & -- & -- \\
		19	& h4r\_ev3\_ctrl					& --					& --			& \checkmark	& --			& -- & -- & -- \\
		20	& iop\_core							& --					& --			& \checkmark	& --			& -- & -- & -- \\
		21	& mas\_domestic\_robotics			& \checkmark			& --			& --			& --			& \checkmark & -- & -- \\
		22	& mavros\_controllers				& \checkmark			& --			& --			& --			& \checkmark & -- & -- \\
		23	& micros\_swarm\_framework			& --					& --			& \checkmark	& --			& -- & \checkmark & -- \\
		24	& motoman\_project					& --					& --			& \checkmark	& --			& -- & -- & -- \\
		25	& moveit							& \checkmark			& \checkmark	& \checkmark	& \checkmark	& \checkmark & \checkmark & (\checkmark )\\
		26	& moveit2							& \checkmark			& \checkmark	& \checkmark	& \checkmark	& \checkmark & \checkmark & (\checkmark )\\
		27	& multimaster\_fkie					& \checkmark			& \checkmark	& --			& \checkmark	& \checkmark & -- & -- \\
		28	& multi\_tracker					& \checkmark			& --			& --			& --			& -- & -- & -- \\
		29	& navigation						& \checkmark			& --			& \checkmark	& --			& \checkmark & \checkmark & -- \\
		30	& neonavigation						& \checkmark			& --			& --			& --			& \checkmark & -- & -- \\
		31	& opencv\_apps						& \checkmark			& \checkmark	& \checkmark	& --			& \checkmark & -- & -- \\
		32	& rapp-platform						& \checkmark			& --			& \checkmark	& \checkmark	& \checkmark & -- & -- \\
		33	& roboracing-software				& \checkmark			& \checkmark	& \checkmark	& (\checkmark )	& \checkmark & (\checkmark ) & -- \\
		34	& rocon\_multimaster				& --					& --			& --			& \checkmark	& -- & -- & -- \\
		35	& ros\_control						& --					& --			& \checkmark	& --			& -- & -- & \checkmark \\
		36	& ros\_people\_model				& \checkmark			& --			& --			& --			& \checkmark & -- & -- \\
		37	& rosplane							& \checkmark			& --			& --			& --			& \checkmark & -- & -- \\
		38	& ros\_tms							& --					& \checkmark	& \checkmark	& \checkmark	& -- & -- & -- \\
		39	& rtabmap\_ros						& --					& \checkmark	& \checkmark	& --			& \checkmark & -- & -- \\
		40	& rtmros\_common					& \checkmark			& --			& --			& --			& \checkmark & -- & -- \\
		41	& SailBoatROS						& --					& --			& --			& \checkmark	& -- & -- & -- \\
		42	& self-driving-golf-cart			& \checkmark			& \checkmark	& \checkmark	& --			& \checkmark & -- & -- \\
		43	& spencer\_people\_tracking			& \checkmark			& \checkmark	& \checkmark	& \checkmark	& -- & -- & -- \\
		44	& sphero\_swarm\_ros				& \checkmark			& --			& --			& --			& \checkmark & -- & -- \\
		45	& teb\_local\_planner				& \checkmark			& --			& \checkmark	& --			& \checkmark & -- & -- \\
		46	& tuw\_multi\_robot					& \checkmark			& \checkmark	& \checkmark	& --			& \checkmark & -- & -- \\
		47	& utexas-art-ros-pkg				& \checkmark			& --			& --			& --			& \checkmark & -- & -- \\
		48	& vigir\_footstep\_planning\_core	& --					& --			& \checkmark	& --			& -- & -- & -- \\
		\bottomrule
	\end{tabular}
\end{smaller}
\end{table*}

	To get a better overview of these \mbox{(sub-)sys}\-tems, we extracted their ROS versions.
	We identified 104 \mbox{(sub-)sys}\-tems that are based on ROS and 10 \mbox{(sub-)sys}\-tems based on ROS2 in the dataset of \cite{Malavolta2021}.
	After filtering these (sub-systems) as described above, only \emph{moveit2} remained as ROS2-based \mbox{(sub-)sys}\-tems that potentially implements reconfiguration.

We then studied the selected (sub-)systems in-depth, analyzing how reconfiguration is implemented (RQ3).
For each match location, we first checked whether the match could be part of a reconfiguration or not, e.g., because the match was in a comment.
Thereafter, we started our in-depth investigation from each code line in which one of the libraries is used.
First, we inspected each usage in detail and identified the purpose for which the library is used at that location.
Afterwards, we did both a forward navigation along the dependencies, i.e., method calls and field accesses, and looked at all the code that was referenced from that location, as well as a backward navigation.
This allowed us to determine how the reconfiguration was triggered and executed.

\subsubsection{Observations}
\begin{figure}
	\includegraphics[width=\columnwidth]{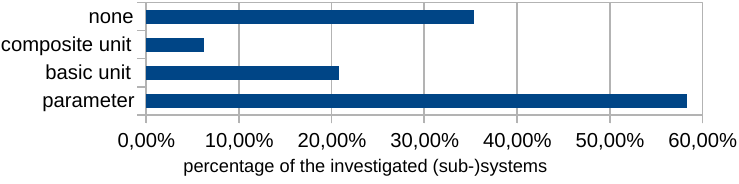}
	\caption{Granularity of reconfiguration implemented in the investigated ROS (sub-)systems}
	\label{fig:applications}
\end{figure}

\Cref{fig:applications} summarizes how often we found which kind of reconfiguration in the investigated robotic \mbox{(sub-)sys}\-tems.
Of the 48 (sub-)systems investigated in-depth, 17 systems contain no reconfiguration at all.
The libraries are used purely for launching the (sub-)systems and sometimes for launch-time configuration.
\Cref{tab:applications} shows in detail which (sub-)systems contain which kind of reconfiguration.
In what follows, we describe implementations of reconfiguration we found.

\begin{lstlisting}[float,
basicstyle=\scriptsize\sffamily,
captionpos=b,
frame=tb,
keepspaces=true,
numbers=left,
numbersep=3pt,
numberstyle=\tiny\color{gray},
tabsize=2,
showstringspaces=false,
showspaces=false,
columns=flexible,
breaklines=true,
aboveskip=2em,
belowskip=1em,
breakatwhitespace=true,
basewidth = {.2em},
breaklines,
language=C++,
caption={Parameter Reconfiguration in ``\emph{AutoRally}''},
label=lst:param]
void CameraTrigger::onInit() {
	...
	// set up dynamic_reconfigure server
	dynamic_reconfigure::Server<camera_trigger_paramsConfig> ::CallbackType cb;
	cb = boost::bind(&CameraTrigger::configCallback, this, _1, _2);
	m_dynReconfigServer.setCallback(cb);
	...
}
...
void CameraTrigger::configCallback(const camera_trigger_paramsConfig &config, uint32_t) {
	m_triggerFPS = config.camera_trigger_frequency;

	// send new FPS to arduino
	m_port.lock();
	m_port.writePort("#fps:" + std::to_string(m_triggerFPS) + "\r\n");
	m_port.unlock();
}
...
\end{lstlisting}

The majority (28 robotics (sub-)systems) 
uses parameter reconfiguration, mostly to reconfigure low-level parameters close to the hardware.
For example, \cref{lst:param} shows a code excerpt of the ``\emph{AutoRally}'' implementation in which parameter reconfiguration is used to update the frequency of a camera.
In lines 2 to 5, the dynamic\_reconfigure server is configured by registering a call back method at the server.
This method is shown in lines 10 to 17 and simply changes the frequency of the camera and writes it to the hardware.
We found no \mbox{(sub-)system} in which parameter values are changed from within the implementation, but these are provided for assignment by external entities.
Surprisingly, only a few \mbox{(sub-)sys}\-tems check assigned parameter values, e.g., whether these are within a plausible range.
While most \mbox{(sub-)sys}\-tems assign the updated parameters immediately to local variables, we still frequently observed locks or flags indicating changes to avoid changing ongoing executions.
For example, in ``\emph{evapi\_ros},'' the configuration of controllers is only updated when the configuration values are changed.
	\Cref{lst:param:flags} shows the implementation of this behavior.
	Whenever the callback method registered with dynamic\_reconfigure is called, a flag is set in line 3.
	During the main execution loop in lines 9 through 17, this flag is checked and only if there are changes, the individual controller configurations are updated.

\begin{lstlisting}[float,
	basicstyle=\scriptsize\sffamily,
	captionpos=b,
	frame=tb,
	keepspaces=true,
	numbers=left,
	numbersep=3pt,
	numberstyle=\tiny\color{gray},
	tabsize=2,
	showstringspaces=false,
	showspaces=false,
	columns=flexible,
	breaklines=true,
	aboveskip=2em,
	belowskip=1em,
	breakatwhitespace=true,
	basewidth = {.2em},
	breaklines,
	language=C++,
	caption={Use of Update Flags to Indicate Changes during Parameter Reconfiguration in ``\emph{avapi\_ros}''},
	label=lst:param:flags]
// Callback function to change pid parameters at runtime.
void CallbackReconfigure(evarobot_controller::ParamsConfig &config, uint32_t level) {
	b_is_received_params = true;
	g_d_wheel_separation = config.wheelSeparation;
	... // copying all remaining values of the config
}

int main(int argc, char **argv) {
	while ( ros::ok() ) {
		// If new parameters are set, ... and controller parameters are updated.
		if ( b_is_received_params ) {
			controller1.UpdateParams(g_d_p_1, g_d_i_1, g_d_d_1, g_d_w_1);
			controller2.UpdateParams(g_d_p_2, g_d_i_2, g_d_d_2, g_d_w_2);
			b_is_received_params = false;
		}
		...
	}
	...
}
\end{lstlisting}

\begin{lstlisting}[float,
	basicstyle=\scriptsize\sffamily,
	captionpos=b,
	frame=tb,
	keepspaces=true,
	numbers=left,
	numbersep=3pt,
	numberstyle=\tiny\color{gray},
	tabsize=2,
	showstringspaces=false,
	showspaces=false,
	columns=flexible,
	breaklines=true,
	aboveskip=2em,
	belowskip=1em,
	breakatwhitespace=true,
	basewidth = {.2em},
	breaklines,
	language=C++,
	caption={Reconfiguration of Basic Units in ``\emph{micros\_swarm\_framework}''},
	label=lst:basic]
AppManager::AppManager():app_loader_("micros_swarm", "micros_swarm::Application") {
	ros::NodeHandle nh;
	app_load_srv_ = nh.advertiseService("app_loader_load_app", &AppManager::loadService, this);
	app_unload_srv_ = nh.advertiseService("app_loader_unload_app", &AppManager::unloadService, this);
	...
}

bool AppManager::loadService(app_loader::AppLoad::Request &req, app_loader::AppLoad::Response &resp) {
	std::string app_name = req.name;
	std::string app_type = req.type;

	bool app_exist = recordExist(app_name);
	if(app_exist) {
		...
		return false;
	}
	else {
		boost::shared_ptr<micros_swarm::Application> app;
		try {
			app = app_loader_.createInstance(app_type);
		}
		catch(pluginlib::PluginlibException& ex) {
			ROS_ERROR(...);
		}
		...
		return true;
	}
}

\end{lstlisting}

\looseness=-1
Nine (sub-)systems contain reconfiguration at the level of basic units.
Functionalities are dynamically enabled, which is usually done by reacting to parameter reconfiguration or providing services.
For example, as shown in \cref{lst:basic}, ``\emph{micros\_swarm\_framework}'' provides services for loading and unloading apps, which are tightly coupled to the framework and, therefore, are classified as basic units by us.
In line 3, the method for loading services, which is defined in lines 8 to 28, is advertised as a ROS service.
When this service is called, this method will load an app using \emph{pluginlib} (line 20) if it is not already running.
However, only two (sub-)systems dynamically load plugins or nodelets (the realization of basic units in ROS) as part of the reconfiguration.
The other system (\emph{cob\_environment\_perception}\footnote{\url{https://github.com/ipa320/cob_environment_perception/blob/indigo_dev/cob_3d_registration/ros/src/registration_nodelet.cpp}}) creates algorithm objects configured for their intended use case, but classloading takes place already at the responsible node's instantiation.
The rest enables or disables basic units by using variables that are checked in conditions to decide whether the basic unit, e.g., a functionality, should be executed or not.
For example, \cref{lst:basic:toggle} shows how ``\emph{AutoRalley}'' allows us to write messages to the topic ``\emph{/runstop}'' to enable motion.
During initialization, the corresponding topic is subscribed (line 3) and a callback function registered, which writes all received messages into class-scope variable.
During execution, this information is processed to decide whether the basic unit implementing movement is enabled or disabled (lines 8--19).

\begin{lstlisting}[float,
	basicstyle=\scriptsize\sffamily,
	captionpos=b,
	frame=tb,
	keepspaces=true,
	numbers=left,
	numbersep=3pt,
	numberstyle=\tiny\color{gray},
	tabsize=2,
	showstringspaces=false,
	showspaces=false,
	columns=flexible,
	breaklines=true,
	aboveskip=2em,
	belowskip=1em,
	breakatwhitespace=true,
	basewidth = {.2em},
	breaklines,
	language=C++,
	caption={Reconfiguration of Basic Units in ``\emph{AutoRalley}'' based on Variables to Enable or Discable Units},
	label=lst:basic:toggle]
void AutoRallyChassis::onInit() {
	...
	runstopSub_ = nh.subscribe("/runstop", 5, &AutoRallyChassis::runstopCallback, this);
	...
}

void runstopCallback(const autorally_msgs::runstopConstPtr& msg) {
	runstops_[msg->sender] = *msg
}

void AutoRallyChassis::setChassisActuators(const ros::TimerEvent&) {
	...
	//check if motion is enabled (all runstop messages = true)
	if(runstops_.empty()) {
		chassisState->runstopMotionEnabled = false;
	} else {
		chassisState->runstopMotionEnabled = true;
		int validRunstopCount = 0;
		for(auto& runstop : runstops_) {
			...
		}
	}
}
\end{lstlisting}

Only three (sub-)systems provide reconfiguration on the level of composite units to some limited extent.
The supported reconfiguration is far behind what is considered in the literature and provided by frameworks such as RobMoSys.
Further, two of these subsystems are ultimately the same subsystem (\emph{moveit}), in its two implementations for ROS and ROS2, respectively.
The other subsystem is \emph{ros$\_$control} that provides a service to load controllers using \textit{pluginlib}.
In principle, this allows structural reconfiguration, but the logic has to be implemented in another subsystem using \emph{ros$\_$control}.
The corresponding code is essentially the same as the one shown in \cref{lst:basic}, with the exception of independently executable composite units being loaded.

\subsubsection{Best Practices for Reconfiguration}
From our insights on the implementations of reconfiguration in robotics (sub-)systems, we derived best practices for implementing reconfiguration.
Since only parameter reconfiguration is widely implemented, we focus on best practices for implementing parameter reconfiguration.

\paragraph
{Patterns for Parameter Reconfiguration.}
Any robotic (sub-)systems that includes parameter reconfiguration must implement its reconfiguration logic.
Depending on how many different parts of the implementation need to be reconfigured to react to a parameter reconfiguration, we observed two best practices for implementing the reconfiguration logic.
To implement an easily comprehensible reconfiguration logic, it is essential to choose the identified practice that best suits the robotic system.

\looseness=-1
The first one is called {\em Reconfigure callback}, reconfiguration logic is implemented in a callback method that is registered at the ROS API \textit{dynamic\_re\-con\-figure}.
In particular, for simple reconfigurations that can be applied immediately to the robotic system, we observed many examples in which this is a simple, but perfectly suitable practice.
The example in \cref{lst:param} can be seen as one instance.
However, for more complex systems, particularly, systems in which reconfiguration has to be applied in multiple parts, following this pattern does not scale.

\looseness=-1
The second pattern addresses the cases in which the first one does not scale. It is called {\em  Message-based} and concerns letting single working parts of the implementation subscribe a topic and to broadcast the new values. It is used when many different and probably independently working parts are affected by a parameter reconfiguration.
Using this practice, the logic specific to an individual part of the robotic system can be located in this part, avoiding overly complex implementations of the callbacks.


\paragraph
{Processing of Updated Values.}
It is essential to be able to update parameter values in the running system without having a negative impact on the current execution.
We mostly observed two ways to implement the processing of updated parameter values: {\em Stateless execution} and {\em stateful execution}.
Both serve as best practices for specific task characteristics as outlined in what follows.

In stateless execution, the reconfigured parameters are applied each time the reconfigured functionality is executed.
Typically, the functionality is executed in a loop, and the parameter values are copied to variables that are in the scope of the loop at the beginning of each iteration. This way there is no interference during execution and the implementation is simple and easily comprehensible.
However, the values are only updated at the beginning of an iteration, making this practice suitable for frequently executed, short-running tasks.

\begin{lstlisting}[float,
	basicstyle=\scriptsize\sffamily,
	captionpos=b,
	frame=tb,
	keepspaces=true,
	numbers=left,
	numbersep=3pt,
	numberstyle=\tiny\color{gray},
	tabsize=2,
	showstringspaces=false,
	showspaces=false,
	columns=flexible,
	breaklines=true,
	aboveskip=2em,
	belowskip=1em,
	breakatwhitespace=true,
	basewidth = {.2em},
	breaklines,
	language=C++,
	caption={Lock-based Parameter Reconfiguration in ``\emph{neonavigation}''},
	label=lst:param:lock]
boost::recursive_mutex parameter_server_mutex_;

// The callback method registered with dynamic_reconfigure
void cbParameter(const SafetyLimiterConfig& config, const uint32_t) {
	boost::recursive_mutex::scoped_lock lock(parameter_server_mutex_);
	hz_ = config.freq;
	timeout_ = config.cloud_timeout;
	... // setting of further config values
}
\end{lstlisting}

For tasks that need updating parameters also during task execution, another practice that considers the state of the running execution is needed.
In stateful execution, the object whose functionality can be reconfigured using parameter reconfiguration, maintains an internal state besides the parameters. Thereby, the parameters can potentially interact with the internal state of the object, requiring a mechanism to notify the object about changed parameters. In the (sub-)systems, we found three different realizations of this practice:
(i) Change Flag -- the object frequently checks a change flag and, if necessary, the reloading of parameter values is triggered, as shown in \cref{lst:param:flags};
(ii)  Lock -- the object provides direct access to the internal configuration values, and a locking mechanism controls the execution while updating live parameters, such as \emph{neonavigation} realizes it using the C++ boost library as shown in \cref{lst:param:lock}; and
(iii)  Callback -- a callback method is provided to stop the current execution and to trigger continuing the execution with the new parameters, such as the implementation of parameter reconfiguration of \emph{rtabmap\_ros} shown in \cref{lst:param:pause} that resumes the camera threat if it has been paused.
Actually starting and stopping the camera is offered via two services.
However, due to the complex execution logic, the latter mechanism is mainly suitable for continuous or long-running tasks.

\begin{lstlisting}[float,
	basicstyle=\scriptsize\sffamily,
	captionpos=b,
	frame=tb,
	keepspaces=true,
	numbers=left,
	numbersep=3pt,
	numberstyle=\tiny\color{gray},
	tabsize=2,
	showstringspaces=false,
	showspaces=false,
	columns=flexible,
	breaklines=true,
	aboveskip=2em,
	belowskip=1em,
	breakatwhitespace=true,
	basewidth = {.2em},
	breaklines,
	language=C++,
	caption={Parameter Reconfiguration in ``\emph{rtabmap\_ros}'' Pausing or Resuming Execution of the Reconfigured Camera If Needed},
	label=lst:param:pause]
class CameraWrapper : public UEventsHandler	{
	// Callback registered at dynamic_reconfigure
	void callback(rtabmap_legacy::CameraConfig &config, uint32_t level) {}
		if(camera) {
			camera->setParameters(config.device_id, config.frame_rate, config.video_or_images_path, config.pause);
		}
	}

	void setParameters(int deviceId, double frameRate, const std::string & path, bool pause) {
		if(cameraThread_) {
			rtabmap::CameraImages * imagesCam = dynamic_cast<rtabmap::CameraImages *>(camera_);
			...
			imagesCam->setImageRate(frameRate);
			if(pause && !cameraThread_->isPaused())	{
				cameraThread_->join(true);
			} else if(!pause && cameraThread_->isPaused()) {
				cameraThread_->start();
			} else { ... }
			...
		}
	}
}
\end{lstlisting}

\paragraph
{Soundness Check of New Values.}
\looseness=-1
To avoid faulty reconfigurations, it is essential to check new, potentially externally provided, values for validity.
Although this is a well-known best practice~\citep{OWASP2021}, we only rarely found such checks in the parameter reconfiguration implemented in the robotic (sub-)systems we studied.
Given the importance of sound parameter values for safe execution, this best practice is related to the validation of reconfigurations that we considered in the SLR.
Since the frameworks do not provide support for checking the validity of reconfiguration, this must be implemented in the robotic (sub-)system itself.
Here, we observed two aspects in the robotic (sub-)systems that need to be considered for properly implementing soundness checks concerning the individual characteristics of a robotic system, namely {\em Time of Check} and {\em Checked Properties}.

\looseness=-1
The new values assigned to unit parameters may be checked at different times, depending on the execution characteristics of the functionalities that use these parameters.  We have observed two common practices: (i) immediate checking -- new parameter values are checked immediately by simple runtime checks to prevent them from being inadvertently processed unchecked;
and (ii) on-demand checking -- if the new parameter values are not used immediately, the checks are performed on demand to avoid unnecessary checks.
On-demand checking is particularly appropriate when there is only one entity that reads them.
Depending on the likelihood that parameters will change again between reconfiguration and the next use of that parameter, immediate checks or on-demand checks may be more suitable.
For unit parameters that are reconfigured more often than they are used, on-demand checks are more suitable, while frequently used but infrequently reconfigured unit parameters can be checked more efficiently with immediate checks.

\looseness=-1
One challenge in implementing soundness checks is to determine what to check. While the properties to check can be application-specific, we identified two practices that should be mostly applicable and that are frequently checked in the (sub-)systems implementing soundness checks of new values: 
(i) value range -- it should be always checked if the new parameter values are within the expected value range; and (ii) consistency -- when having multiple parameters, it is essential to check if these are consistent and expected relations among the unit parameters are fulfilled; the same might apply if the parameter value has to relate to the current system state.

	\subsection{Discussion of Reconfiguration Mechanisms}
	The majority of the literature focuses on approaches for specifying structural reconfiguration using domain-specific languages (DSLs) and executing them at runtime.
	During this execution, constraints on the validity of target configurations are usually checked or only valid configurations are computed by design.
	In contrast, robotic frameworks only provide low-level application programming interfaces (APIs) for loading or unloading assets and changing parameter values.
	Consequently, the necessary logic for implementing the structural reconfiguration considered in the literature must be handwritten by the developers of robotic systems.
	Academic robotic frameworks can provide wrappers around more low-level frameworks that allow specifying the reconfiguration logic using models.
	However, in robotic (sub-)systems, we did not observe such reconfiguration, but only the frequent use of parameter reconfiguration.
	A particular challenge arises from ensuring the validity of reconfigurations and avoiding invalid target configurations and or intermediate states. While the frameworks provide no support in this sense, we identified concrete practices for implementing reconfiguration safely.
	Further academic work, such as GenoM \citep{1997Genome}, specifically focuses on generating code with corresponding validations.
	However, this work was not covered by our literature review and not observed in practice.

\section{Discussion}
\label{sec:discussion}

We now discuss our results. In particular, we relate the outcomes of our three data sources, the SLR, the frameworks, and the robotic (sub-)systems. But, we also discuss our observations with respect to the state-of-the-art of reconfiguration in other domains.

\subsection{State-of-the-Art vs State-of-Practice}

From the analysis of the state-of-the-art as captured by our SLR, we identified several techniques to structural reconfiguration.
As shown in Figure~\ref{fig:slr:granularity}, the most popular granularity for reconfiguration is composite unit, followed by unit parameter, and basic unit.
Thereby, the configurations have primarily a lifespan at the scope of a task or mission (see \cref{fig:slr:lifespan}).
However, the frameworks do not provide advanced support in terms of reconfiguration triggers and logic for this kind of reconfiguration, only low-level APIs that allow manual implementation of structural reconfiguration.
Probably related to this limitation, in robotics, ROS-based (sub-)systems, as shown in \Cref{fig:applications}, structural reconfiguration is almost completely absent, and only parameter reconfiguration is used intensively.
The low-level APIs provided by ROS are primarily used for implementing parameter reconfiguration.
Thereby, the best practices concerning safety  derived by us are only applied in few robotics \mbox{(sub-)systems.}

One explanation for this discrepancy could be that these reconfigurations are not necessary in real-world settings and in non-academic setups.
Another explanation could be that the required advanced reconfiguration support is only now entering the robotics domain, e.g., in projects such as RobMoSys, but is not widely available.
Also, current robotic systems have not yet reached the complexity envisioned by the scientific community.
However, we expect them to do so in the near future, and this will trigger the need to provide sophisticated reconfiguration support to robotic system designers.
We believe that robots will increasingly become multi-purpose,  i.e., able to perform various tasks, with the consequent need of customization and reconfiguration capabilities.  Moreover, they will be increasingly required to be used in uncontrolled environments, often shared with humans. This is visible also in the various competitions and challenges organized in robotics conferences and forums. The demand for autonomy, run-time reconfiguration, and deployment of software also after production, as well as the use of AI solutions, will often need to deal with requirements of compliance to security and safety standards. This is indeed another dimension of complexity that the community will need to face; the compliance to safety and security standards will become incremental and continuous as it is already happening in the automotive domain \citep{continuousCompliance}. Finally, the programming and use of robots should become more accessible to end-users without knowledge in robotics or ICT. This will be a need both for industrial and service robots. In fact, according to our experience, small and medium enterprises are sometimes reluctant to use robotics solutions for automating their tasks because they miss the capabilities to program and configure robots according to their (evolving) needs.

In the related domain of autonomous driving, we already found such an example of composite unit-level reconfiguration in the open-source driving system Autoware.auto~\citep{autoware}.
In this system, different motion planners are used based on the current driving scenario, such as lane following, lane changing, or parking.
The reconfiguration mechanism is a custom implementation of this project, and they use behavior trees~\citep{colledanchise2018behavior} as an executable model for specifying the reconfiguration logic.
Considering the assumed complexity of reconfiguration in the reviewed literature, this example is still relatively simple and is the only example of coarse-grained reconfiguration that we found in this system, but may indicate a future need for more complex reconfiguration.

Another driver for reconfiguration in robotics could be machine learning (ML).
Today, many ML-enabled system, i.e., robotic systems, employ complex data processing pipelines involving multiple ML models, e.g., for multi-stage perception of the environment \citep{Peldszus2024a}.
In this context, particularly due to the size of models and the numerous tasks implemented in robotic systems using ML, i.e., the number of ML models used.
Particularly large language models (LLM), such as GPT, Llama, Claude, or Qwen, need multiple gigabytes of memory to be executed.
For example, we measured that a Qwen2.5-72b LLM \citep{Yang2024} with 3.5 bits per word and 24k tokens context length already needs 40GB of GPU memory.
Due to this huge memory demands it is infeasible to execute all models in parallel but it may be necessary to reconfigure the robotic system, i.e., which model is loaded.
This could be a main driver for structural reconfiguration at the granularity of composite units or even base units in practice.
Also, this could lead to resource limitations, which we observed in our SLR to be not being a main reason for reconfiguration any longer, to become more relevant again.

	One of the obstacles to the more widespread adoption of reconfiguration may be the additional complexity it introduces and the necessity to address it in all phases of development.
	Among other things, this added complexity is particularly challenging for testing.
	In a previous interview study we conducted to validate a reactive security monitor capable of reconfiguring the monitored system \citep{Peldszus2024}, e.g., to put it into a safe mode in case of detected malicious actions, a security engineer from a large automotive supplier pointed out the testing challenges.
	Standards such as SOTIF, which can be relevant to any robotic system, require companies to test all possible operating modes for safety, which can be challenging in the context of reconfiguration, since all possible configurations must be tested in combination with all points in time at which reconfigurations can occur.

\subsection{Reconfiguration in Other Domains}
\label{sec:related}
The reconfiguration mechanisms provided by robotic frameworks, especially the non-academic ones, are currently limited.
However, state-of-the-art frameworks in many domains allow sophisticated reconfiguration, e.g.,
the Linux kernel, the Debian Linux distribution, the Eclipse IDE (OSGi), and the Android operating system~\citep{berger.ea:2014:ecosystems}.
Others, such as the eCos operating system, only support compile-time configuration, but come with sophisticated DSLs and configuration tools that may be suitable for reconfiguration.

The investigated robotic frameworks have a well-defined asset base, but with the exception of RobMoSys, they lack a clear specification of the reconfiguration space.
The Linux kernel and eCos can serve as inspiration with their feature-model-like DSLs to specify the configuration space.
Robotic systems would benefit from explicit configuration space modeling 
also for the reconfiguration at runtime, e.g., 
to analyse possible configurations and potential conflicts that could occur at runtime~\citep{patrick2021seip} or to analyse security vulnerabilities~\citep{Peldszus2018}.
Positively, with the RobMoSys project, researchers have already been working on providing such methods and tools to the robotic domain.

Robotic frameworks provide more interfaces for interaction among the assets than state-of-the-art software ecosystems~\citep{berger.ea:2014:ecosystems} 
that mainly rely on the programming language-specific interface specifications and focus on direct source code interactions among the assets.
The focus on pure source code interfaces in the state-of-the-art software ecosystems allows mainly static and dynamic linking as interaction mechanisms.
Only Eclipse and Android, which support sophisticated class or module loading, need an interaction manager at runtime.
Such a manager or even multiple managers are provided by all robotic frameworks.
In the end, the interaction management in the robotic frameworks is comparable to Eclipse's services and extension points.

When it comes to the reconfiguration mechanisms, the software ecosystems are comparable to the investigated robotic frameworks.
In the Linux kernel, kernel modules are dynamically loaded when these are accessed.
The same applies to plugins in Eclipse, which can be dynamically loaded by OSGi once any of the contained Java classes is accessed.
Like most robotic frameworks, none of the software ecosystems provides techniques for explicitly specifying reconfiguration.
However, unlike the robotic frameworks, reconfiguration is also not considered one of the core aspects in them.
\section{Threats to Validity}
	A threat to the validity of the SLR could arise from the fact that the initial paper selection depends primarily on the second author.
	To mitigate this threat, we tended to consider a paper as potentially relevant rather than discarding it.
	In addition, we performed a snowballing iteration that allowed important but initially erroneously discarded papers to be reconsidered.
	In fact, of the 5,089 initially excluded papers, 48 papers were reconsidered in this way, and we decided to look at 24 of them in more detail using multiple authors, but only included 6 papers for classification.

	The focus on ROS (sub-)systems in the review of how reconfiguration is implemented in practice represents a threat to external validity.
	However, ROS could be considered representative because it is the only widely used robotic middleware.
	Although our investigation of robotic frameworks has revealed that ROS and even ROS2 do not provide advanced reconfiguration techniques, our investigation shows that reconfiguration is actually implemented in ROS systems.
	Therefore, ROS might not be the perfect example for investigating reconfiguration in practice but we also lack better practically adopted frameworks, which yet have to be developed.

	Since ROS2-based systems are underrepresented in the sample of robotic systems studied, our results apply only to ROS-based systems.
	However, we did not find any major improvements in ROS2 regarding reconfiguration during our framework review, nor did we find any differences between \emph{moveit} and \emph{moveit2} regarding reconfiguration.
	This suggests that our findings may also apply to ROS2-based systems, but it remains for future work to confirm this hypothesis.

A further threat is that we only considered open-source systems and we did not consider bespoke or industrial frameworks. In a follow-up study, it would be interesting to investigate also these frameworks since they could have better reconfiguration mechanisms and/or make use of a custom middleware.

A possible source of bias arises from the backgrounds of the authors and might threaten internal validity.
We mitigated this bias by including a diverse set of authors, including authors from the software engineering domain, and authors whose expertise is in robotics.
Moreover, as a template basis for our comparison, we used the categorization from \cite{berger.ea:2014:ecosystems}.
Two related threats are that this template might give mainly a software viewpoint, as opposed to a hardware viewpoint, and that it might be outdated.
For mitigation, we adapted the template based on our SLR and secondary literature.

Finally, while we provided justification for our conceptual model based on the papers considered in our SLR, we did not perform additional evaluation, independent of the SLR, for it.
Such an evaluation, e.g., based on expert opinions, could provide further valuable insight.
However, we are confident that the conceptual model in its present form serves its purpose for this paper, to compare how available state-of-practice frameworks implement the reconfiguration of robots described in the state-of-the-art literature.
\begin{table}
	\caption{Summarized answers to our RQs}
	\label{tab:summaryRQ}
	\begin{tabular}{p{8.4cm}}
		{\bf Answers to the research questions}\\
		\hline
		\vspace{.01em}
		{\em RQ1: What are the motivations for developing dynamically reconfigurable robotic software systems?}\\
		Researchers consider reconfiguration of robotic systems mostly to deal with changes in dynamic environments and due to the execution of various tasks by a single robot.
		Further, the reconfiguration as measure for reacting to hardware or software faults is also a popular motivation.
		Less common motivations relate to limited resources and to react to safety issues.
		Finally, reconfiguration is frequently used as synonym for other concepts, such as, deployment-time variability.\smallskip\\
		\hline
		\vspace{.01em}
		{\em RQ2: What aspects of a robotic system can be reconfigured?}\\
		While literature mainly focuses on the structural reconfiguration of robotic systems,
		robotic frameworks support this granularity only to a limited extent and require developers to implement the entire reconfiguration logic.
		Academic robotic frameworks can provide wrappers around more low-level frameworks that allow specifying the reconfiguration logic using models.
		So far, only parameter reconfiguration has been widely used in robotic (sub-)systems.\smallskip\\
		\hline
		\vspace{.01em}
		{\em RQ3: What mechanisms are used for developing dynamically reconfigurable robotic software systems?}\\
		\looseness=-1
		The literature mainly focuses on approaches for specifying structural reconfiguration using DSLs and executing them 
		at runtime.
		Instead, robotic frameworks provide low-level APIs for loading or unloading assets and changing parameter values.
		The 
		logic for implementing the structural reconfiguration considered in the literature has to be handwritten by the developers of robotic systems.
		However, in robotic (sub-)systems, we did not observe such reconfiguration, but only the frequent use of parameter reconfiguration. 
 \smallskip\\
		\hline
	\end{tabular}
\end{table}

\begin{table}
	\caption{Implications for researchers and practitioners}
	\label{tab:summaryImplications}
	\begin{tabular}{p{8.4cm}}
		{\bf Implications}\\
		\hline
		\vspace{.01em}
		{\em Academics:} In the summary of RQ1 above, we highlight that reacting to safety issues is a not so common motivation for developing dynamically reconfigurable robotic software systems. However, since often robots need to work in collaboration with humans and/or in safety critical domains, there is the need of rigorous software enginering approaches in robotics.
		Due to the identified significant mismatch between the state-of-the-art and state-of-practice, it is essential to further collaborate with practitioners. On one side, this is necessary to guarantee that academic research is aligned with industrial needs, and, on the other side, to enable technology transfer from academia to industry.\smallskip\\
		\hline
		\vspace{.01em}
		{\em Practitioners:} Robots will be multi-purpose and 
		need to deal with high levels of uncertainty, e.g.,  in the environment, reflected in sensor performance and reliability.
		This will probably lead to more complexity and variability in robots and, consequently, requires structural and more sophisticated reconfiguration techniques. It will become increasingly important to decouple the specification of reconfigurability from the application logic. Available DSLs we surveyed in the state-of-the-art could become relevant. Moreover, explicitly specifying the configuration space of robotic systems allows configuration tools and configuration editors, and increases the reliability of reconfigurable robotic systems.
		The need of
structural and more sophisticated reconfiguration techniques may require to mitigate the identified discrepancies between the state-of-the-art and state-of-practice concerning reconfiguration.  Robotic frameworks could be extended to (i) explicitly support context reconfiguration, (ii) provide support for reconfiguration triggers and logic, taking inspiration from academic frameworks that provide more sophisticated languages for specifying reconfiguration, and (iii) provide more advanced APIs than those low-level provided by ROS so to enable derived best practices concerning safety.
\smallskip\\
		\hline
	\end{tabular}
\end{table}
\section{Conclusion}
\label{sec:conclusion}

\noindent
We determined the state-of-the-art and the state-of-practice of \edit{automated runtime} software reconfiguration in robotics.
We analyzed the former by surveying the literature on reconfiguration in robotic systems, inspecting \edit{78 relevant papers} in detail.
We analyzed the latter by reviewing how four major robotic frameworks support reconfiguration and how reconfiguration is realized in 48 real robotic \mbox{(sub-)sys}\-tems.
Based on our analysis of robotic \mbox{(sub-)sys}\-tems, we derived best practices for implementing parameter reconfiguration. 
Table~\ref{tab:summaryRQ} provides an answer to the research questions and Table~\ref{tab:summaryImplications} draws  take away messages for both academics and practitioners.
We identified a significant mismatch between the state-of-the-art and state-of-practice in reconfiguration of robotic systems.
As future work, we plan to further investigate the discrepancies we identified to unleash their reasons and to identify research directions and strategies to fill this gap.
This mainly concerns structural and more sophisticated reconfiguration approaches that we believe will be needed in robotics in the near future.
Since the observed discrepancies might already stem from the motivation for reconfiguration, we will follow up on the motivations for reconfigurations for which until now we only captured the academic perspective by interviewing developers of robotic systems.

\section{Data Availability Statements}
Our replication package provides all raw data, scripts that have been used, and results~\citep{replication}.

\section{Acknowledgments}
We thank Sergio García for discussions on earlier versions of this paper.

This work has been partially funded by the European Union - NextGenerationEU under the Italian Ministry of University and Research (MUR) National Innovation Ecosystem grant ECS00000041 - VITALITY – CUP: D13C21000430001.
This work has been also partially funded by the European Union - NextGenerationEU under the Italian Ministry of University and Research (MUR) National Innovation Ecosystem, grant PE0000020 – CHANGES – CUP: D53C22002560006
This work has been also partially funded by the European Union - NextGenerationEU under the Italian Ministry of University and Research (MUR) National Innovation Ecosystem, grant PE00000014 – SERICS – CUP: D33C22001300002.
Also, the authors acknowledge the support of the MUR (Italy) (i) Department of Excellence 2023 - 2027, (ii) PRIN PNRR 2022 project ``RoboChor: Robot Choreography'' (grant P2022RSW5W), and (iii) PRIN project ``HALO: etHical-aware AdjustabLe autOnomous systems'' (grant 2022JKA4SL).


\bibliographystyle{spbasic}
\bibliography{IEEEabrv,doc}

\end{document}